\documentclass[11pt]{article}

\usepackage[preprint]{acl}

\usepackage{times}
\usepackage{latexsym}

\usepackage[T1]{fontenc}

\usepackage[utf8]{inputenc}

\usepackage{microtype}

\usepackage{inconsolata}

\usepackage{graphicx}
\usepackage{amsmath} 
\usepackage{booktabs}
\usepackage{multirow}
\usepackage{listings}
\usepackage{enumitem}
\newcommand{\tref}[1]{Table~\ref{#1}}
\newcommand{\fref}[1]{Figure~\ref{#1}}

\newcommand{\apref}[1]{Appendix~\ref{#1}}

\newcommand{\sref}[1]{Section~\ref{#1}}
\newcommand{\OURDATA}{\textbf{\texttt{HybridDeepResearch}}}

\title{Benchmarking Hybrid Deep Research Across Database Querying and Web Search}

\author{\textbf{
Ruofan Wu\textsuperscript{\textmd 1}\thanks{~Equal contribution.}
\quad
Peiran Xu\textsuperscript{\textmd 2}\footnotemark[1]
\quad
Xiaolong Li\textsuperscript{\textmd 3}\footnotemark[1]
\quad
Fan Shu\textsuperscript{\textmd 1}\footnotemark[1]
}\\
\textbf{
Soyoung Yoon\textsuperscript{\textmd 4}
\quad
Yite Wang\textsuperscript{\textmd 5}\thanks{~Project lead.}
\quad
Xiaodong Yu\textsuperscript{\textmd 5}\footnotemark[2]
\quad
Boyi Liu\textsuperscript{\textmd 5}
}\\
\textbf{
Feng Yan\textsuperscript{\textmd 1}
\quad
Debiao Li\textsuperscript{\textmd 2}
\quad
Yuxiong He\textsuperscript{\textmd 5}
\quad
Zhewei Yao\textsuperscript{\textmd 5}
}\\
{}\textsuperscript{1}University of Houston \quad
{}\textsuperscript{2}University of California, Los Angeles \quad
{}\textsuperscript{3}The University of Hong Kong \\
{}\textsuperscript{4}Seoul National University \quad
{}\textsuperscript{5}Snowflake AI Research \\
}

\begin{document}
\maketitle

\begin{abstract}

While autonomous agents have made significant strides in ``deep research'' by iteratively navigating the open web to synthesize information, real-world problem-solving is rarely confined to a single environment. Complex analytical tasks inherently require agents to weave together evidence from both ambiguous unstructured text (e.g., the open web) and highly precise structured data (e.g., relational databases). However, existing benchmarks evaluate these modalities in isolation, failing to capture the critical ``handoff'' — the ability to preserve constraints when moving evidence between systems. We introduce \OURDATA{}, to our knowledge the first deep-research benchmark that requires both web search and SQL to form a complete, verifiable answer. The benchmark contains 380 tool-dependent tasks grounded in LiveSQLBench-Base-Lite databases and public web corpora, validated through automated checks and human review, and covering three reasoning patterns: \texttt{SQL2S}, \texttt{S2SQL}, and \texttt{Parallel}. Evaluations across proprietary and open-weight models under various agentic scaffolds reveal that even state-of-the-art models like GLM-5.2, Claude-Sonnet-4.6 and GPT-5 achieve only about 50–54\% Pass@8 on the hard subset. Notably, results show that directional reasoning is substantially more difficult than parallel intersection, highlighting that bridging structured and unstructured information spaces without losing constraints remains a major open challenge for agentic systems. Code and datasets are publicly available at \href{https://github.com/Snowflake-AI-Research/HybridDeepResearch}{GitHub} and \href{https://huggingface.co/datasets/Snowflake/HybridDeepResearch}{Hugging Face}.

\end{abstract}

\begin{figure*}[t]
    \centering
    \includegraphics[width=0.8\textwidth]{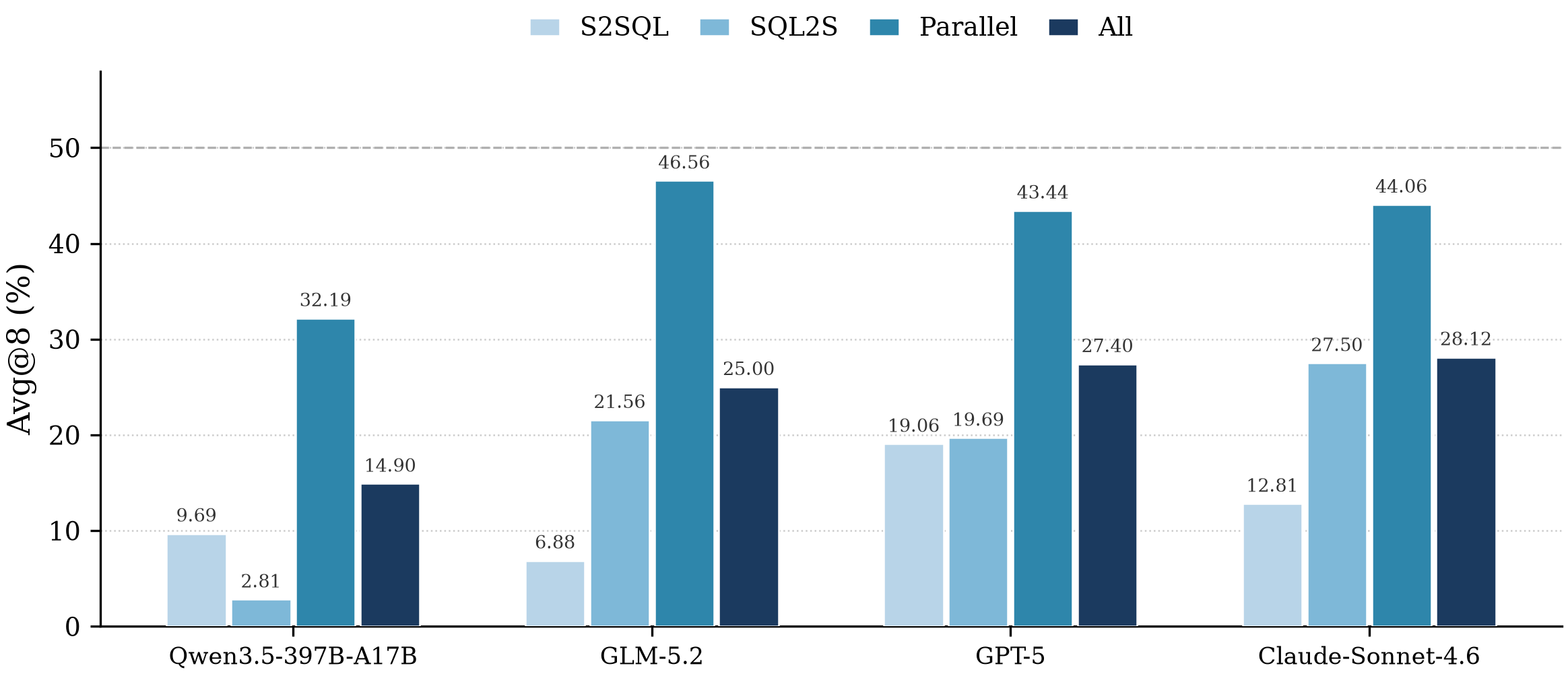}
    \caption{
    Avg@8 performance of representative frontier and open-weight models on the balanced hard subset.
    Overall success remains below $30\%$, with substantially lower performance on the directional reasoning patterns.
    }
    \label{fig:hard-subset-avg8}
\end{figure*}
\begin{figure*}[h]
\centering
\includegraphics[width=\textwidth]{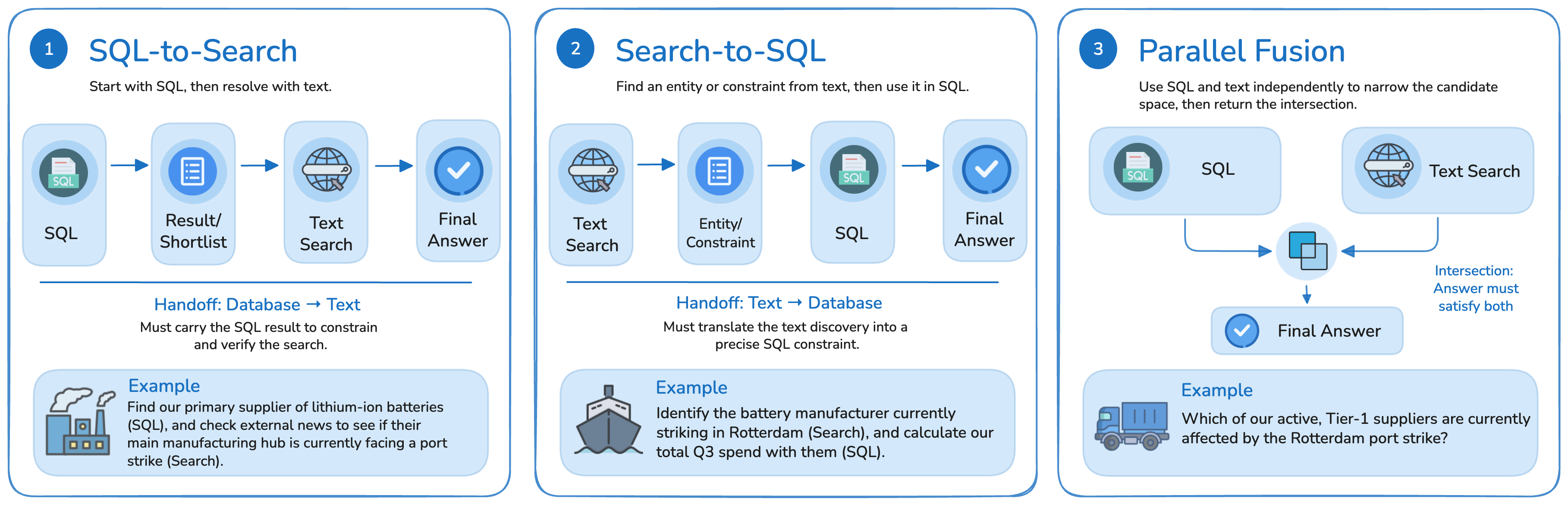}
\caption{Three hybrid reasoning patterns in \OURDATA{} and the information flow required by each.
\texttt{SQL2S} passes a database-recovered entity to web search, whereas \texttt{S2SQL} converts a search-recovered entity or constraint into a SQL predicate.
\texttt{Parallel} solves the SQL and search components independently and returns their set intersection.
Arrows indicate the direction in which evidence or constraints must be transferred between tools.}
\label{fig:hybrid-patterns}
\end{figure*}

\section{Introduction}


While large language models have achieved remarkable success in navigating dynamic web environments~\cite{gou2026mind2web} and synthesizing evidence from open corpora~\cite{wei2025browsecomp,chen2025browsecompplus,du2026deepresearch}, these achievements primarily reflect single-modality reasoning. In complex real-world analytical tasks, however, critical information does not reside in a vacuum. High-stakes domains like financial auditing, supply-chain risk assessment, and scientific research inherently depend on integrating disparate data sources. To answer real-world questions, an autonomous agent must traverse the boundary between highly structured data (such as relational databases) and ambiguous, unstructured information (such as news articles, public filings, or the open web). Despite this clear need, the evaluation landscape remains heavily compartmentalized. Foundational models are typically assessed on their ability to either write complex SQL queries~\cite{li2023can,livesqlbench2025,lei2025spider} or perform open-ended web retrieval independently~\cite{zheng2025deepresearcher,java2026characterizing,li2025webthinker}. Consequently, an agent might achieve state-of-the-art performance in isolated text-to-SQL or browsing tasks, yet completely break down when required to coordinate evidence across both. We identify this critical failure mode as the ``handoff.'' When evidence must be transferred, cross-checked, and validated across systems, agents frequently lose track of constraints.

To make these orchestration failures visible and quantifiable, we introduce \OURDATA{}, a benchmark explicitly designed to act as a diagnostic tool for hybrid deep research. Rather than measuring isolated tool use, every task in our framework necessitates a verifiable intersection of modalities. We structure these tasks around three distinct reasoning patterns, each designed to expose specific handoff vulnerabilities:
\begin{itemize}[noitemsep, topsep=2pt, leftmargin=*]
    \item \textbf{\texttt{SQL2S} (SQL-to-Search):} The database provides a rigid anchor that must constrain subsequent web exploration.
    \item \textbf{\texttt{S2SQL} (Search-to-SQL):} Unstructured web search uncovers an entity or condition that must dictate a follow-up database query.
    \item \textbf{\texttt{Parallel}:} SQL and web search independently produce candidate sets, and the final answer emerges only from their exact intersection.
\end{itemize}
\OURDATA{} features $380$ complex questions grounded in LiveSQLBench-Base-Lite databases~\cite{livesqlbench2025} alongside public evidence drawn from Wikipedia and FineWeb-10BT~\cite{penedo2024fineweb}. Every task is strictly tool-dependent, requiring successful execution of both SQL queries and web searches. For focused stress-testing, we curate a balanced $120$-example hard subset containing exactly $40$ uniquely challenging questions for each of the three reasoning patterns.


We establish comprehensive baselines using state-of-the-art proprietary models (including GPT-5 and Claude-Sonnet-4.6) and open-weight models (such as Qwen3.5 and GLM-5.2), deployed under both single-agent and multi-agent frameworks~\cite{smolagents,su2026miroflow}.
On the balanced hard subset, even the most capable proprietary systems reach only about $50$--$54\%$ Pass@8 and remain below $30\%$ overall Avg@8, indicating that successful solutions are neither frequent nor reliable. As shown in \fref{fig:hard-subset-avg8}, the largest performance gaps occur on the directional handoffs, \texttt{SQL2S} and \texttt{S2SQL}.
These results demonstrate that preserving constraints across disparate data silos remains a major unsolved challenge for deep-research agents.



In summary, our key contributions are:
\begin{itemize}
    \item We introduce \OURDATA{}, to our knowledge the first deep-research evaluation framework explicitly designed to measure the intersection of open-ended web search and structured database querying.
    \item We define three reasoning patterns, \texttt{SQL2S}, \texttt{S2SQL}, and \texttt{Parallel}, and construct $380$ automatically validated and human-reviewed, tool-necessary questions with a balanced $120$-example hard subset. 
    \item We conduct extensive evaluations across leading models and agentic scaffolds, revealing a critical performance bottleneck in how modern agents transfer and reconcile evidence between distinct modalities.
\end{itemize}

\section{Related Work}
Prior work relevant to \OURDATA{} spans deep-research evaluation, database-agent evaluation, and hybrid benchmarks. Deep-research benchmarks focus on complex questions that require iterative web search and reasoning over retrieved evidence \cite{wei2025browsecomp,java2026characterizing,chen2025browsecompplus}. Database evaluation has progressed from mapping natural-language questions to executable SQL to agent-oriented tasks that require multi-step interaction with database environments \cite{yu2018spider,li2023can,lei2025spider,huo2026bird}.  Existing hybrid benchmarks combine heterogeneous evidence \cite{chen2020hybridqa,chen2021ottqa,zhu2021tatqa,christmann2024compmix}, but do not explicitly evaluate answer-oriented tasks in which web search and executable database querying are both necessary under different dependency structures. \OURDATA{} targets this setting through SQL-to-Search, Search-to-SQL, and Parallel tasks. A detailed discussion is provided in \apref{app:related-work}.
\begin{figure*}[h]
\centering
\includegraphics[width=\textwidth]{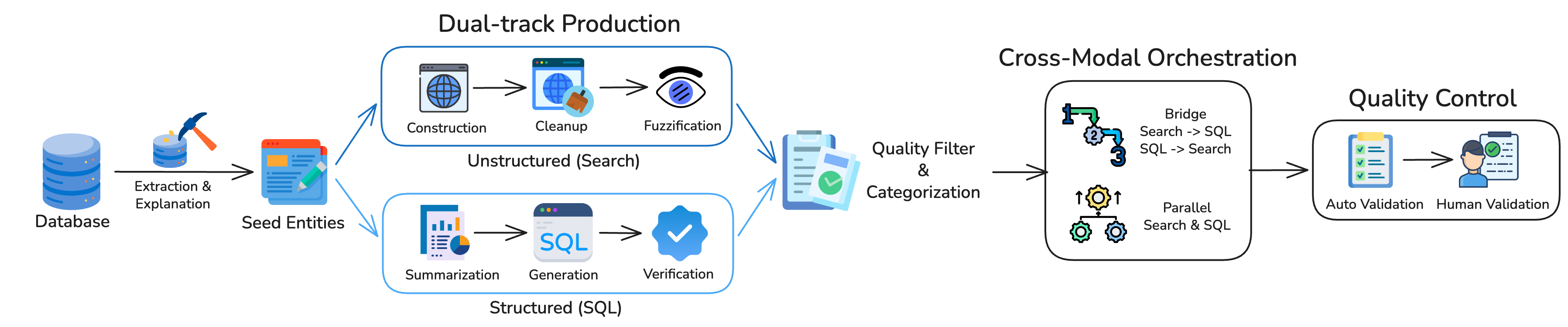}
\caption{Overview of the \OURDATA{} construction pipeline.
Database values are grounded as shared seed entities and developed along separate tracks into evidence-backed search questions and executable SQL question--query pairs.
After track-specific filtering, role-compatible components are composed into \texttt{S2SQL}, \texttt{SQL2S}, or \texttt{Parallel} instances.
Automatic single-modality checks and final human review remove items that are solvable without both tools, ambiguous, or non-unique.}
\label{fig:pipeline}
\end{figure*}

\section{Hybrid Reasoning Patterns}
\label{sec:philosophy}


Real-world hybrid research is challenging not only because it requires two tools, but because the constraints passing between them are often structurally complex, ambiguous, and susceptible to severe distraction. To systematically stress-test agents, we categorize these cross-modal handoffs into three reasoning patterns (illustrated in \fref{fig:hybrid-patterns}). We intentionally design these patterns to expose the gap between isolated tool proficiency and true cross-modal orchestration.

\textbf{\texttt{SQL2S} (SQL-to-Search).}
The structured database provides a rigid anchor that must strictly constrain subsequent unstructured web exploration. For example, an agent may be asked to find the primary supplier of lithium-ion batteries and check external news to see if their main manufacturing hub faces a port strike. The agent must first query the database to identify the specific supplier (e.g., Company A), and then issue a targeted web search for Company A's strike status. A common failure mode is the ``salience trap,'' where the agent successfully finds Company A via SQL, but issues a generic web search for battery strikes and erroneously returns a highly ranked news article about a different company.

\textbf{\texttt{S2SQL} (Search-to-SQL).}
Unstructured web search uncovers an entity or condition that must dictate a follow-up database query. For instance, if asked to identify the battery manufacturer striking in Rotterdam and calculate Q3 spend with them, the agent must first search the news to identify the striking company (Company A). It must then translate that textual discovery into a strict SQL predicate (e.g., \texttt{WHERE vendor\_name = `Company A'}). A frequent failure is ``dropping the anchor,'' where the agent identifies the correct company from text but writes a generic SQL query that fails to apply the necessary filter, thus breaking the cross-modal link.

\textbf{\texttt{Parallel}.}
SQL and web search independently produce candidate sets, and the final answer emerges only from their exact intersection. If asked which active, Tier-1 suppliers are affected by the Rotterdam strike, the agent must query the database for a list of Tier-1 suppliers and independently search the web for companies affected by the strike. The agent succeeds only if it returns the precise overlap. A common failure is premature fusion, where the agent returns all striking companies found on the web, failing to cross-check them against the database's active supplier constraints.

While these patterns differ in how information flows, they test the same underlying capability:
whether an agent can preserve, translate, and reconcile semantic constraints across structurally distinct modalities.
We utilize these patterns as the foundational templates for constructing hybrid tasks that enforce a  verifiable ``hybrid lock''.
\section{\OURDATA{} Construction}
\label{sec:method}

\begin{figure*}[t]
\centering
\includegraphics[width=\textwidth]{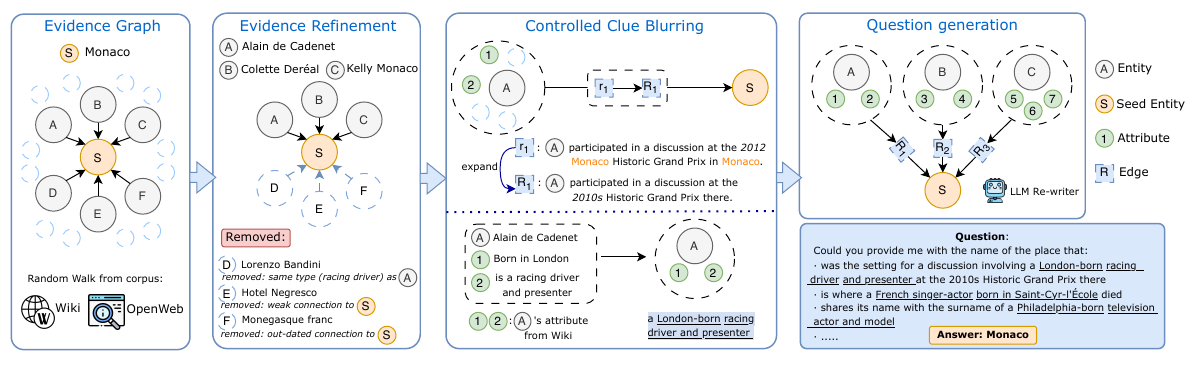}
\caption{Construction of the unstructured search component from a database grounded seed entity.
Public evidence from Wikidata, Wikipedia, and FineWeb is organized into a convergent multi-clue graph, then refined to remove ambiguous, weak, or outdated relations.
Controlled clue blurring suppresses direct entity leakage before an LLM renders the verified graph as a role-compatible search question for \texttt{SQL2S}, \texttt{S2SQL}, or \texttt{Parallel}.}
\label{fig:search-pipeline-example}
\end{figure*}

Constructing a robust hybrid deep research benchmark requires more than a naive pairing of a database query and a web-retrieval task. Each benchmark instance must satisfy four coupled requirements: (1) both components must be anchored to a shared, real-world entity, (2) remain individually verifiable, (3) instantiate one of the cross-modal dependency structures defined in \sref{sec:philosophy}, and (4) strictly necessitate the synthesis of both modalities to derive the final answer. Violating any of these conditions yields questions that are either ambiguous, trivially solvable via a single source, or hybrid merely in appearance.

To satisfy these constraints, we design a rigorous construction pipeline centered around three core mechanisms: \textit{Dual-Grounded Entity Anchoring}, \textit{Role-Aware Symmetric Generation}, and \textit{The Hybrid Lock} (summarized in \fref{fig:pipeline}).

\subsection{Dual-Grounded Entity Anchoring}
\label{sec:seed}

To establish a reliable bridge between structured and unstructured environments, we anchor each task to a shared seed entity. However, discovering valid bridge entities is challenging: database values are often rigid but suffer from naming ambiguities (e.g., ``Monaco'' could denote a country, a city, or a racing event), while the open web contains a vast, noisy entity space.

To resolve this, we employ a dual-grounding mechanism. We first extract database cell values denoting concrete entities (e.g., people, organizations) and validate their global salience using the Google Knowledge Graph. Crucially, global salience does not guarantee local database consistency. We therefore deploy a SQL-capable exploration agent to examine how each retained entity is utilized within its specific schema context, generating a \emph{database-grounded description}. This description isolates the precise real-world meaning of the entity, guiding subsequent cross-modal generation to retrieve evidence that strictly matches the database context rather than unrelated homonyms.

\begin{figure*}[t]
\centering
\includegraphics[width=\textwidth]{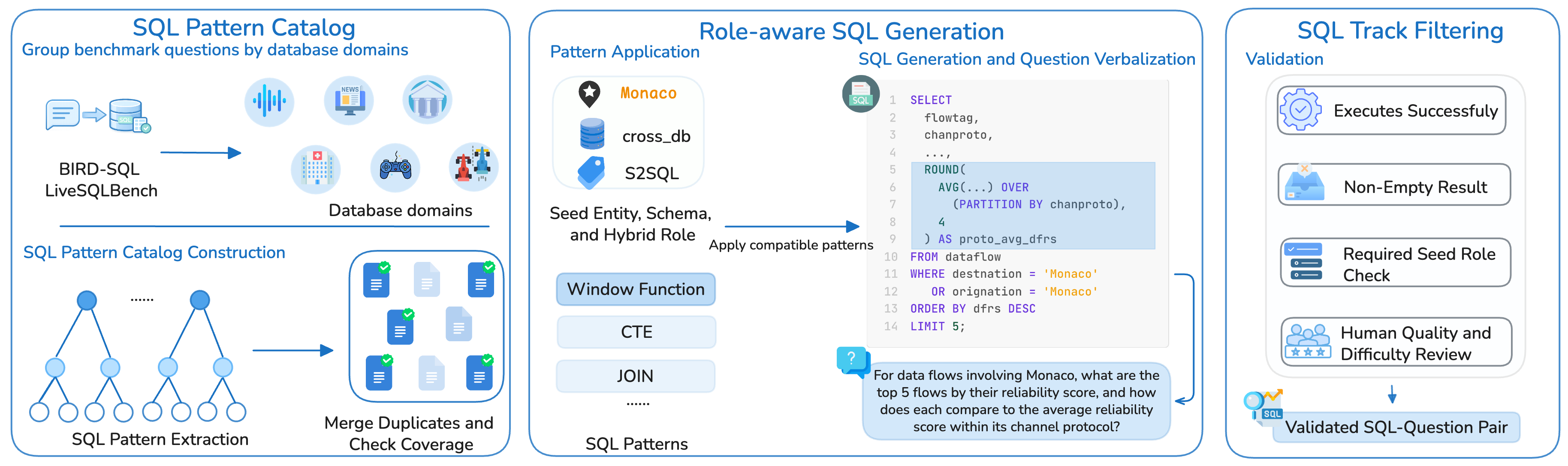}
\caption{Construction of the structured SQL component from a database-grounded seed entity.
Reusable SQL patterns are extracted from BIRD-SQL and LiveSQLBench queries across database domains, then deduplicated and checked for coverage.
Each compatible pattern is instantiated under the target hybrid role, producing SQL that is verbalized as a natural-language question and retained after execution, seed-role, and human-quality checks.}
\label{fig:sql-pipeline-example}
\end{figure*}

\subsection{Role-Aware Symmetric Generation}

Rather than independently sampling questions, we symmetrically generate search and SQL components tailored to the seed's required role in the target handoff. The two tracks are generated independently but are bound by the shared seed.

\paragraph{Unstructured Track: Search Questions.} 
The web-search component requires agents to synthesize public evidence without relying on parametric memory or trivial keyword matching. As illustrated in \fref{fig:search-pipeline-example}, we construct these questions through four phases:
\begin{itemize}[noitemsep, topsep=2pt, leftmargin=*]
    \item \textbf{Evidence Graph Construction:} We gather public evidence from different sources---Wikidata property triples~\cite{vrandecic2014wikidata} and Wikipedia--- as well as long-tail web documents from FineWeb-10BT~\cite{penedo2024fineweb}. We organize converging clues into multi-hop graphs (e.g., $c_1 \rightarrow a \leftarrow c_2$).
    \item \textbf{Evidence Refinement:} We rigorously prune weak, ambiguous, or outdated relations from the graph, ensuring every edge expresses a concrete, verifiable relation.
    \item \textbf{Controlled Clue Blurring:} To prevent single-clue shortcuts, entity names within the clues are abstracted into non-identifying descriptive attributes (e.g., generalizing ``2012'' to ``the 2010s''), and relations are paraphrased.
    \item \textbf{Question Generation:} An LLM rewrites the blurred graph into a natural-language question using inverted-question templates, strictly tailoring the phrasing to match the seed's role in the target hybrid pattern (e.g., phrasing it as a provided premise for \texttt{SQL2S}, or the unknown target for \texttt{S2SQL}).
\end{itemize}

\paragraph{Structured Track: Role-Constrained SQL.}
The SQL component pairs a natural-language question with an executable query on LiveSQLBench-Base-Lite~\cite{livesqlbench2025}. Unconstrained LLM generation typically yields repetitive, shallow queries. As illustrated in \fref{fig:sql-pipeline-example}, we enforce structural complexity in two steps:
\begin{itemize}[noitemsep, topsep=2pt, leftmargin=*]
    \item \textbf{SQL Pattern Catalog:} We curate a reusable catalog of 35 advanced SQL patterns (e.g., CTE aggregations followed by joins) organized into 6 categories and three difficulty levels, derived from human-written queries in BIRD-SQL~\cite{li2023can} and LiveSQLBench. These patterns were validated by human reviewers to ensure structural coverage and validity.
    \item \textbf{Role-Aware Generation:} We apply compatible patterns to the seed entity based on its schema. For \texttt{S2SQL}, the seed is generated as a strict predicate (e.g., a \texttt{WHERE} condition). For \texttt{SQL2S}, the query returns a singleton output containing the seed. For \texttt{Parallel}, the query returns a multi-row candidate set containing the seed.
\end{itemize}

\subsection{Cross-Modal Composition and The ``Hybrid Lock''}

\begin{figure}[t]
\centering
\includegraphics[width=0.5\textwidth]{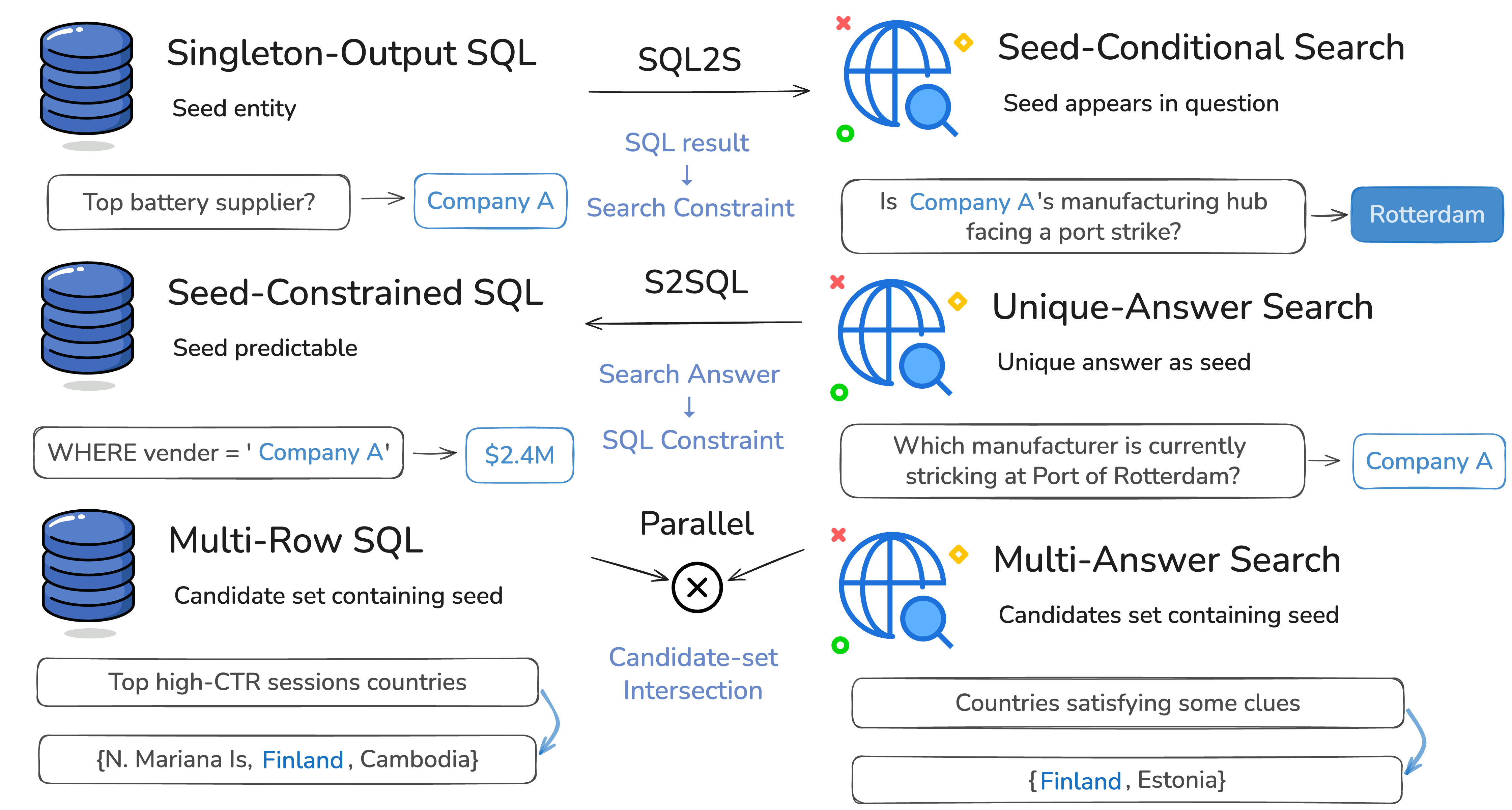}
\caption{Role-compatible composition of independently generated SQL and search components.
In \texttt{SQL2S}, a singleton SQL result constrains the subsequent search; in \texttt{S2SQL}, a unique search answer becomes a SQL predicate; and in \texttt{Parallel}, the two candidate sets are intersected.
The examples illustrate how the shared seed is transferred or matched so that both modalities are necessary to obtain the final answer.}
\label{fig:composition-example}
\end{figure}

After independently filtering both tracks for answerability and execution correctness (detailed in Appendix~\ref{app:filtering}), we orchestrate the compatible components to form complete hybrid tasks. 

As shown in \fref{fig:composition-example}, we employ \emph{directional composition} for \texttt{S2SQL} and \texttt{SQL2S}, where the exact output of one modality becomes the rigid constraint passed into the other. For \texttt{Parallel}, we employ \emph{set intersection}, pairing a multi-answer search component with a multi-row SQL component, retaining only pairs whose intersection uniquely identifies the shared entity.

\paragraph{Enforcing the Hybrid Lock.}
The most critical vulnerability in hybrid benchmark construction is accidental single-source leakage. To test whether both tools are required, every final question undergoes a single-tool ablation. An evaluator agent attempts to solve each item twice: once utilizing only web search, and once utilizing only SQL execution. Any item that can be solved by a single modality—whether through parametric guessing, query leakage, or overly broad search hits—is immediately discarded. This final quality control step reduces common shortcuts and empirically strengthens the ``hybrid lock,'' such that the final questions require successful coordination of web search and SQL under the evaluated settings. Finally, human reviewers perform a concluding pass to eliminate any remaining ambiguous or poorly phrased questions.
Full details of the human review process are provided in \apref{app:human-review}.
\section{\OURDATA{} Statistics}
\label{sec:statistics}

\begin{figure}[t]
\centering
\includegraphics[width=0.5\textwidth]{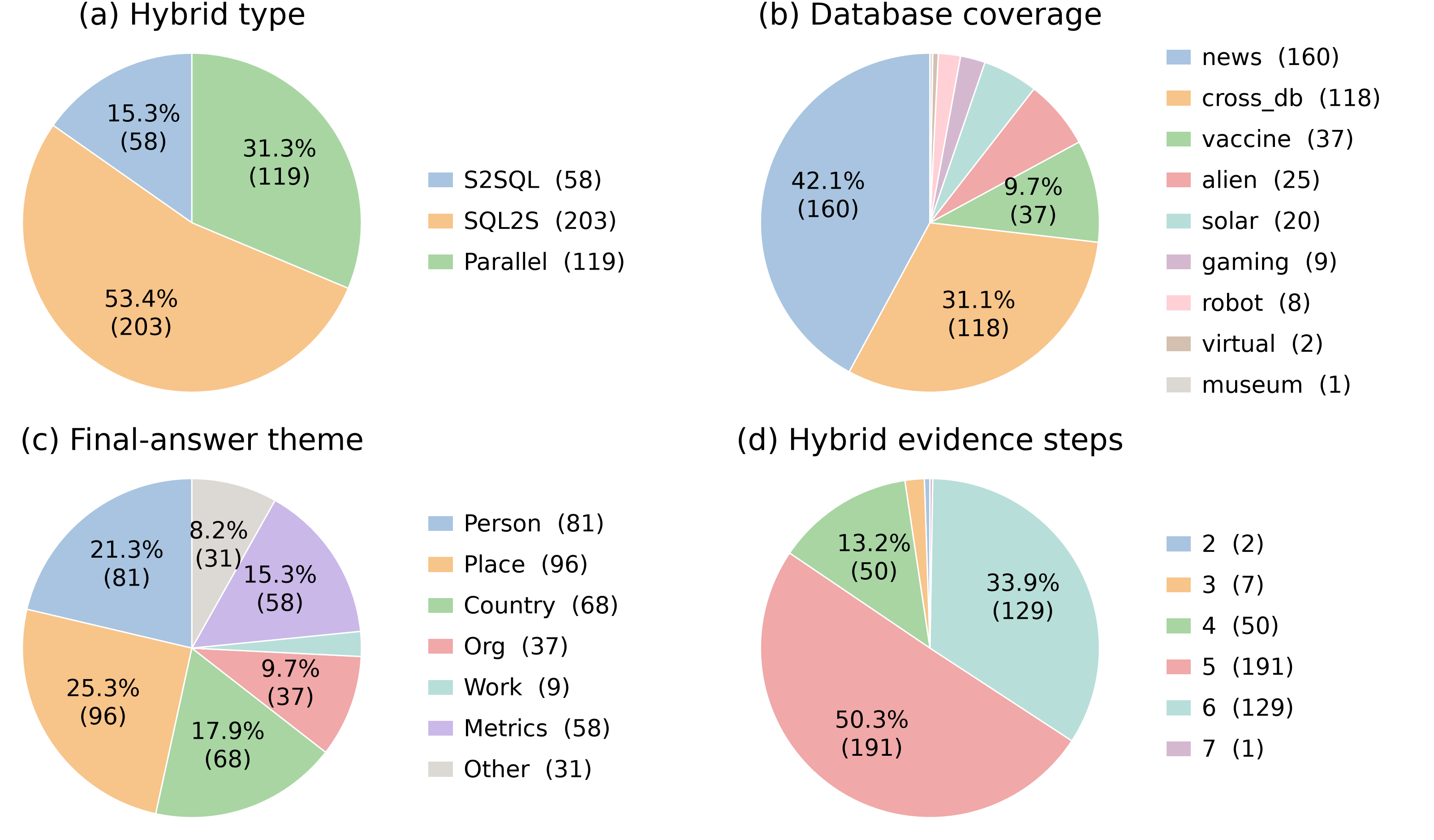}
\caption{
Distribution of hybrid types, source databases, final-answer themes, and evidence steps in the full \OURDATA{} set.
}
\label{fig:dataset-stats}
\end{figure}

The full \OURDATA{} set contains $380$ hybrid questions grounded in nine LiveSQLBench-Base-Lite databases.
\fref{fig:dataset-stats} summarizes the dataset by hybrid type, source database, answer category, and evidence depth.

\paragraph{Dataset composition.}
By hybrid type, \texttt{SQL2S} is the largest subset with $203$ questions, followed by \texttt{Parallel} with $119$ and \texttt{S2SQL} with $58$.
Questions are concentrated in the \texttt{news} and \texttt{cross\_db} databases, which together account for most questions, with the remaining examples spread across seven smaller domains.
Final answers cover people, places, countries or regions, organizations, works, and numerical or tabular results returned by SQL.
Evidence depth, defined as the number of search reasoning steps plus one SQL step, is concentrated at five and six steps.
\apref{app:examples} provides one complete example for each hybrid type.

\paragraph{Database knowledge and hints.}
Each question is associated with the database knowledge needed for SQL execution through \texttt{sql\_knowledge\_id}.
Most questions use between one and four knowledge snippets.
A small subset of $10$ questions ($2.6\%$) also includes an explicit \emph{hint}.
These hints define task-specific rules that cannot be recovered from the database schema alone, such as score thresholds, weighted formulas, or exclusion criteria.

\paragraph{Balanced hard subset.}
Because the full set is imbalanced across hybrid types and repeated evaluation can be costly, we also construct a balanced subset of $120$ questions, with $40$ examples from each type.
The subset was manually selected from the full set to emphasize the more challenging questions and is used for controlled comparisons under limited API budgets. 
\section{Evaluation}
\label{sec:evaluation}

\subsection{Experimental Setup}
\label{sec:exp-setup}

We evaluate Qwen3.5~\cite{qwen3.5}, GLM-5.2~\cite{zeng2026glm}, GPT-5~\cite{openai2025gpt5}, and Claude-Sonnet-4.6~\cite{anthropic2026sonnet46} under two agent frameworks, smolagents~\cite{smolagents} and MiroFlow~\cite{su2026miroflow}.
Open-weight models are evaluated on the full benchmark, while proprietary models are evaluated on the balanced hard subset with smolagents.
We compare Real Web with Corpus Search, a fixed-corpus retrieval backend.
Full implementation and inference details are provided in \apref{app:exp-details}.

For \texttt{S2SQL}, we execute the predicted and gold SQL queries and compare their results.
For \texttt{SQL2S} and \texttt{Parallel}, we use an LLM judge to compare the predicted and gold answers.
Each task is evaluated over eight trials.
Pass@8~\cite{chen2021evaluating} measures whether the task is solved at least once, while Avg@8 reports the average success rate.

\begin{table}[t]
\centering
\resizebox{\columnwidth}{!}{%
\begin{tabular}{llcc}
\toprule
\textbf{Scaffold} & \textbf{Search} & \textbf{Pass@8} & \textbf{Avg@8} \\
\midrule
\multirow{2}{*}{smolagents}
 & Real Web & 58.95 & \textbf{32.80} \\
 & Corpus Search & 56.58 & 32.43 \\
\midrule
\multirow{2}{*}{MiroFlow}
 & Real Web & \textbf{66.32} & 32.43 \\
 & Corpus Search & 63.42 & 32.76 \\
\bottomrule
\end{tabular}%
}
\caption{Effect of agent scaffold and retrieval backend on the full \OURDATA{} benchmark ($380$ instances), using Qwen3.5-397B-A17B-FP8 as the shared backbone.
Each task is run eight times.
Results are overall scores aggregated across the three hybrid types and bold denotes the best value in each metric column.}
\label{tab:scaffold-search}
\end{table}

\begin{table*}[t]
\centering
\small
\resizebox{0.95\textwidth}{!}{%
\begin{tabular}{llcccccccc}
\toprule
 & & \multicolumn{4}{c}{\textbf{smolagents}} & \multicolumn{4}{c}{\textbf{MiroFlow}} \\
\cmidrule(lr){3-6} \cmidrule(lr){7-10}
\textbf{Model} & \textbf{Metric} & All & S2SQL & SQL2S & Parallel & All & S2SQL & SQL2S & Parallel \\
\midrule
\multirow{2}{*}{Qwen3.5-4B}
 & Pass@8 & 36.05 & 12.07 & 21.67 & 72.27 & 41.58 & 17.24 & 29.56 & 73.95 \\
 & Avg@8 & 17.20 & 3.23 & 8.19 & 39.39 & 15.79 & 2.37 & 11.02 & 30.46 \\
\midrule
\multirow{2}{*}{Qwen3.5-9B}
 & Pass@8 & 40.53 & 17.24 & 21.18 & 84.87 & 53.16 & 25.86 & 41.38 & 86.55 \\
 & Avg@8 & 20.59 & 4.53 & 10.41 & 45.80 & 22.34 & 4.96 & 17.30 & 39.39 \\
\midrule
\multirow{2}{*}{Qwen3.5-27B}
 & Pass@8 & 52.89 & 39.66 & 35.96 & 88.24 & 57.11 & 43.10 & 48.77 & 78.15 \\
 & Avg@8 & 26.74 & 15.73 & 15.15 & 51.89 & 23.75 & 17.24 & 21.06 & 31.51 \\
\midrule
\multirow{2}{*}{Qwen3.5-122B-A10B}
 & Pass@8 & 43.95 & 27.59 & 30.54 & 74.79 & 53.95 & 43.10 & 46.80 & 71.43 \\
 & Avg@8 & 20.23 & 9.05 & 13.42 & 37.29 & 24.93 & 18.53 & 22.72 & 31.83 \\
\midrule
\multirow{2}{*}{Qwen3.5-397B-A17B}
 & Pass@8 & 56.58 & \textbf{51.72} & 41.87 & 84.03 & 63.42 & 58.62 & 53.20 & 83.19 \\
 & Avg@8 & 32.43 & \textbf{20.69} & 19.52 & 60.19 & 32.76 & 22.20 & 28.45 & 45.27 \\
\midrule
\multirow{2}{*}{GLM-5.2-FP4}
 & Pass@8 & \textbf{72.37} & 46.55 & \textbf{68.97} & \textbf{90.76} & \textbf{75.79} & \textbf{62.07} & \textbf{69.46} & \textbf{93.28} \\
 & Avg@8 & \textbf{43.52} & 13.15 & \textbf{38.61} & \textbf{66.70} & \textbf{49.24} & \textbf{38.15} & \textbf{43.97} & \textbf{63.66} \\
\bottomrule
\end{tabular}%
}
\caption{Open-weight model performance on the full \OURDATA{} benchmark ($380$ instances) with Corpus Search under the single-agent smolagents and multi-agent MiroFlow scaffolds.
\texttt{All} reports the aggregate score, while \texttt{S2SQL}, \texttt{SQL2S}, and \texttt{Parallel} give type-specific results.
Bold marks the best model within each scaffold, task type, and metric.}
\label{tab:open-full}
\end{table*}

\begin{table*}[t]
\centering
\resizebox{1.3\columnwidth}{!}{%
\begin{tabular}{llcccc}
\toprule
\textbf{Model} & \textbf{Metric} & \textbf{All} & \textbf{S2SQL} & \textbf{SQL2S} & \textbf{Parallel} \\
\midrule
\multirow{2}{*}{Qwen3.5-397B-A17B}
 & Pass@8 & 34.17 & 30.00 & 15.00 & 57.50 \\
 & Avg@8 & 14.90 & 9.69 & 2.81 & 32.19 \\
\midrule
\multirow{2}{*}{GLM-5.2-FP4}
 & Pass@8 & 50.83 & 30.00 & 45.00 & \textbf{77.50} \\
 & Avg@8 & 25.00 & 6.88 & 21.56 & \textbf{46.56} \\
\midrule
\multirow{2}{*}{Claude-Sonnet-4.6}
 & Pass@8 & 52.50 & 32.50 & \textbf{47.50} & \textbf{77.50} \\
 & Avg@8 & \textbf{28.12} & 12.81 & \textbf{27.50} & 44.06 \\
\midrule
\multirow{2}{*}{GPT-5}
 & Pass@8 & \textbf{54.17} & \textbf{45.00} & 40.00 & \textbf{77.50} \\
 & Avg@8 & 27.40 & \textbf{19.06} & 19.69 & 43.44 \\
\bottomrule
\end{tabular}%
}
\caption{Model comparison on the balanced hard subset of \OURDATA{} under smolagents with Corpus Search.
The subset contains $120$ instances, with $40$ each from \texttt{S2SQL}, \texttt{SQL2S}, and \texttt{Parallel}. \texttt{All} therefore averages the three types equally.
Bold marks the best value in each task-type and metric column.}
\label{tab:subset-smol}
\end{table*}

\subsection{Results}
\label{sec:results}

We report results along three dimensions: agent scaffolds and retrieval backends, open-weight model performance across hybrid patterns, and comparison with proprietary models on the balanced hard subset.

\paragraph{RQ1. How do agent scaffolds and retrieval backends affect performance?}
We compare smolagents and MiroFlow under Real Web and Corpus Search on the full set, using Qwen3.5-397B-A17B-FP8 as the shared backbone (\tref{tab:scaffold-search}).
MiroFlow achieves higher Pass@8 under both retrieval backends, increasing the score from $58.95$ to $66.32$ with Real Web and from $56.58$ to $63.42$ with Corpus Search.
However, Avg@8 remains nearly unchanged, suggesting that the multi-agent scaffold increases the chance of solving a question at least once without consistently improving success across repeated trials.

With this backbone, Real Web provides a small Pass@8 advantage over Corpus Search under both scaffolds, while Avg@8 differs only slightly.
This indicates that the fixed-corpus backend remains competitive with live-web retrieval while providing a more controlled evaluation setting.

\paragraph{RQ2. How well do current open-weight models solve the three hybrid reasoning patterns?}
\tref{tab:open-full} reports full-set results for open-weight models under both scaffolds with Corpus Search.
Performance within the Qwen3.5 family does not increase monotonically with total parameter count.
The 397B-A17B model is strongest overall among the Qwen models, but the 27B dense model outperforms the 122B-A10B MoE model in several settings.
Across all matched Qwen sizes, MiroFlow improves Pass@8 over smolagents, with the largest gains generally appearing on the directional patterns, \texttt{S2SQL} and \texttt{SQL2S}.
Its effect on Avg@8 is less consistent, and performance on \texttt{Parallel} sometimes decreases despite a higher overall Pass@8.

GLM-5.2-FP4 is the strongest open-weight model under both scaffolds.
With MiroFlow, it reaches $75.79$ Pass@8 and $49.24$ Avg@8 overall.
Compared with Qwen3.5-397B-A17B, GLM-5.2 shows broad gains across task types, with particularly large improvements in Avg@8 and \texttt{SQL2S} Pass@8.
Nevertheless, the directional patterns remain substantially harder than \texttt{Parallel}.
For example, under MiroFlow, GLM-5.2 obtains $62.07$ Pass@8 on \texttt{S2SQL}, compared with $93.28$ on \texttt{Parallel}.
The large gap between Pass@8 and Avg@8 suggests that many questions are solved only occasionally rather than reliably.

\paragraph{RQ3. How do proprietary models compare with strong open-weight baselines on the balanced hard subset?}
\tref{tab:subset-smol} compares Claude-Sonnet-4.6 and GPT-5 with GLM-5.2-FP4 and Qwen3.5-397B-A17B on the balanced $120$-example hard subset under smolagents with Corpus Search.
GPT-5 achieves the highest overall Pass@8 at $54.17$, while Claude-Sonnet-4.6 achieves the highest overall Avg@8 at $28.12$.
No model dominates across all task types: GLM-5.2 ties for the highest \texttt{Parallel} Pass@8 and obtains the highest \texttt{Parallel} Avg@8.

Even the strongest systems solve only about half of the hard-subset questions under Pass@8.
The directional patterns are again the main bottleneck:
GPT-5 performs best on \texttt{S2SQL}, while Claude performs best on \texttt{SQL2S}, but both remain well below their \texttt{Parallel} scores.

We also evaluate Claude-Sonnet-4.6 and GPT-5 in two trials on the full set, with results reported in \apref{app:full-api}, which provides a noisier estimate.
Claude performs better overall, while GPT-5 remains stronger on \texttt{S2SQL}.
The different type distribution likely contributes to this ranking change:
the hard subset is balanced with $40$ examples per type, whereas the full set contains only $58$ \texttt{S2SQL} questions out of $380$.

\subsection{Analysis}
\label{sec:analysis}


We observe three recurring failure modes.
On \texttt{Parallel} tasks, agents may return a candidate supported by only one source instead of intersecting the SQL and search results.
This often occurs when the agent commits to an early candidate and fails to verify that it also satisfies the constraints from the other tool.
On \texttt{SQL2S} tasks, a correct database result may be lost during subsequent web search.
The agent may issue a broader query that no longer preserves the entity or condition recovered from SQL.
On \texttt{S2SQL} tasks, agents may identify the correct entity but use an incorrect SQL operation or formula.
In these cases, the handoff succeeds at the entity level, but the retrieved information is not translated into the correct database constraint or computation.

The two scaffolds also exhibit a clear trade-off.
MiroFlow's additional worker attempts can recover constraints missed earlier, which helps improve Pass@8.
However, this broader exploration requires more LLM and tool calls and leads to more timeouts.
This is consistent with the fact that MiroFlow improves Pass@8 without consistently improving Avg@8.
Detailed trajectory statistics and case studies are provided in \apref{app:trajectory-analysis}.

\section{Conclusion}

We introduced \OURDATA{}, a benchmark for deep-research agents that must jointly reason over open web evidence and relational databases.
It contains $380$ questions spanning three hybrid task patterns, \texttt{SQL2S}, \texttt{S2SQL}, and \texttt{Parallel}, together with a balanced $120$-example hard subset.

Across proprietary and open-weight models, current agents remain unreliable when information must be passed between web search and SQL.
Multi-agent scaffolds improve Pass@8, but directional tasks remain substantially harder than parallel intersection.
These results identify cross-tool consistency as a key challenge for deep-research agents.
\section*{Limitations}

\OURDATA{} currently uses nine LiveSQLBench-Base-Lite databases together with public web corpora such as Wikipedia and FineWeb-10BT.
This controlled setting does not cover proprietary schemas, specialized scientific databases, or the full diversity of the live web.
We have completed an extension of the construction pipeline on a newly built database, which will support an expanded release and public leaderboard. 

The full set is also imbalanced across hybrid types.
\texttt{SQL2S} is the largest category, while \texttt{S2SQL} contains only $58$ of the $380$ questions.
We provide a balanced $120$-example hard subset to support fairer comparison across types.
Increasing \texttt{S2SQL} coverage remains an important direction for future expansion.

\section*{Acknowledgments}
We would like to thank Nikki Lijing Kuang and Han Wang for insightful discussions and valuable feedback on this work.

\section*{Ethical Considerations}

The benchmark is intended for evaluating research agents rather than deployment in high-stakes settings.
Because it uses public web evidence, it may inherit factual errors, biases, or objectionable content from its sources.
Agent outputs may also contain unsupported claims or incorrect database results and should be independently verified before practical use.
Repeated multi-agent evaluation additionally incurs non-trivial computational cost.

\bibliography{references}

@article{wei2025browsecomp,
  title={BrowseComp: A Simple Yet Challenging Benchmark for Browsing Agents},
  author={Wei, Jason and Sun, Zhiqing and Papay, Spencer and McKinney, Scott and Han, Jeffrey and Fulford, Isa and Chung, Hyung Won and Passos, Alex Tachard and Fedus, William and Glaese, Amelia},
  journal={arXiv preprint arXiv:2504.12516},
  year={2025}
}

@article{chen2025browsecompplus,
  title={BrowseComp-Plus: A More Fair and Transparent Evaluation Benchmark of Deep-Research Agent},
  author={Chen, Zijian and Ma, Xueguang and Zhuang, Shengyao and Nie, Ping and Zou, Kai and Liu, Andrew and Green, Joshua and Patel, Kshama and Meng, Ruoxi and Su, Mingyi and Sharifymoghaddam, Sahel and Li, Yanxi and Hong, Haoran and Shi, Xinyu and Liu, Xuye and Thakur, Nandan and Zhang, Crystina and Gao, Luyu and Chen, Wenhu and Lin, Jimmy},
  journal={arXiv preprint arXiv:2508.06600},
  year={2025}
}

@inproceedings{yang2018hotpotqa,
  title={HotpotQA: A Dataset for Diverse, Explainable Multi-hop Question Answering},
  author={Yang, Zhilin and Qi, Peng and Zhang, Saizheng and Bengio, Yoshua and Cohen, William W. and Salakhutdinov, Ruslan and Manning, Christopher D.},
  booktitle={Proceedings of the 2018 Conference on Empirical Methods in Natural Language Processing},
  pages={2369--2380},
  year={2018},
  doi={10.18653/v1/D18-1259}
}

@article{trivedi2022musique,
  title={MuSiQue: Multihop Questions via Single-hop Question Composition},
  author={Trivedi, Harsh and Balasubramanian, Niranjan and Khot, Tushar and Sabharwal, Ashish},
  journal={Transactions of the Association for Computational Linguistics},
  volume={10},
  pages={539--554},
  year={2022},
  doi={10.1162/tacl_a_00475}
}

@article{vrandecic2014wikidata,
  title={Wikidata: A Free Collaborative Knowledgebase},
  author={Vrande{\v{c}}i{\'c}, Denny and Kr{\"o}tzsch, Markus},
  journal={Communications of the ACM},
  volume={57},
  number={10},
  pages={78--85},
  year={2014},
  doi={10.1145/2629489}
}

@article{penedo2024fineweb,
  title={The FineWeb Datasets: Decanting the Web for the Finest Text Data at Scale},
  author={Penedo, Guilherme and Kydl{\'{i}}{\v{c}}ek, Hynek and Ben Allal, Loubna and Lozhkov, Anton and Mitchell, Margaret and Raffel, Colin and Von Werra, Leandro and Wolf, Thomas},
  journal={arXiv preprint arXiv:2406.17557},
  year={2024}
}

@Misc{smolagents,
  title={smolagents: a smol library to build great agentic systems.},
  author={Aymeric Roucher and Albert Villanova del Moral and Thomas Wolf and Leandro von Werra and Erik Kaunismäki},
  howpublished ={\url{https://github.com/huggingface/smolagents}},
  year={2025}
}

@article{su2026miroflow,
  title={Miroflow: Towards high-performance and robust open-source agent framework for general deep research tasks},
  author={Su, Shiqian and Xing, Sen and Dong, Xuan and Zhong, Muyan and Wang, Bin and Zhu, Xizhou and Chen, Yuntao and Wang, Wenhai and Deng, Yue and Zhu, Pengxiang and others},
  journal={arXiv preprint arXiv:2602.22808},
  year={2026}
}

@article{chen2021evaluating,
  title={Evaluating large language models trained on code},
  author={Chen, Mark and Tworek, Jerry and Jun, Heewoo and Yuan, Qiming and Pinto, Henrique Ponde De Oliveira and Kaplan, Jared and Edwards, Harri and Burda, Yuri and Joseph, Nicholas and Brockman, Greg and others},
  journal={arXiv preprint arXiv:2107.03374},
  year={2021}
}

@inproceedings{kwon2023efficient,
  title={Efficient Memory Management for Large Language Model Serving with PagedAttention},
  author={Woosuk Kwon and Zhuohan Li and Siyuan Zhuang and Ying Sheng and Lianmin Zheng and Cody Hao Yu and Joseph E. Gonzalez and Hao Zhang and Ion Stoica},
  booktitle={Proceedings of the ACM SIGOPS 29th Symposium on Operating Systems Principles},
  year={2023}
}

@misc{qwen3.5,
  title  = {{Qwen3.5}: Towards Native Multimodal Agents},
  author = {{Qwen Team}},
  month  = {February},
  year   = {2026},
  url    = {https://qwen.ai/blog?id=qwen3.5}
}

@article{zeng2026glm,
  title={Glm-5: from vibe coding to agentic engineering},
  author={Zeng, Aohan and Lv, Xin and Hou, Zhenyu and Du, Zhengxiao and Zheng, Qinkai and Chen, Bin and Yin, Da and Ge, Chendi and Huang, Chenghua and Xie, Chengxing and others},
  journal={arXiv preprint arXiv:2602.15763},
  year={2026}
}

@misc{livesqlbench2025,
  author       = {BIRD Team},
  title        = {LiveSQLBench: A Dynamic and Contamination-Free Benchmark for Evaluating LLMs on Real-World Text-to-SQL Tasks},
  year         = {2025},
  howpublished = {https://github.com/bird-bench/livesqlbench},
  note         = {Accessed: 2025-05-22}
}

@inproceedings{yu2018spider,
  title={Spider: A large-scale human-labeled dataset for complex and cross-domain semantic parsing and text-to-sql task},
  author={Yu, Tao and Zhang, Rui and Yang, Kai and Yasunaga, Michihiro and Wang, Dongxu and Li, Zifan and Ma, James and Li, Irene and Yao, Qingning and Roman, Shanelle and others},
  booktitle={Proceedings of the 2018 conference on empirical methods in natural language processing},
  pages={3911--3921},
  year={2018}
}

@article{li2023can,
  title={Can llm already serve as a database interface? a big bench for large-scale database grounded text-to-sqls},
  author={Li, Jinyang and Hui, Binyuan and Qu, Ge and Yang, Jiaxi and Li, Binhua and Li, Bowen and Wang, Bailin and Qin, Bowen and Geng, Ruiying and Huo, Nan and others},
  journal={Advances in Neural Information Processing Systems},
  volume={36},
  pages={42330--42357},
  year={2023}
}

@article{gou2026mind2web,
  title={Mind2web 2: Evaluating agentic search with agent-as-a-judge},
  author={Gou, Boyu and Huang, Zanming and Ning, Yuting and Gu, Yu and Lin, Michael and Qi, Weijian and Kopanev, Andrei and Yu, Botao and Jimenez Gutierrez, Bernal and Shu, Yiheng and others},
  journal={Advances in Neural Information Processing Systems},
  volume={38},
  year={2026}
}

@inproceedings{lei2025spider,
  title={Spider 2.0: Evaluating language models on real-world enterprise text-to-sql workflows},
  author={Lei, Fangyu and Chen, Jixuan and Ye, Yuxiao and Cao, Ruisheng and Shin, Dongchan and Su, Hongjin and Suo, Zhaoqing and Gao, Hongcheng and Hu, Wenjing and Yin, Pengcheng and others},
  booktitle={International Conference on Learning Representations},
  volume={2025},
  pages={28691--28735},
  year={2025}
}

@inproceedings{du2026deepresearch,
  title={Deepresearch bench: A comprehensive benchmark for deep research agents},
  author={Du, Mingxuan and Xu, Benfeng and Zhu, Chiwei and Zhang, Licheng and Wang, Xiaorui and Mao, Zhendong},
  booktitle={International Conference on Learning Representations},
  volume={2026},
  pages={42414--42448},
  year={2026}
}

@article{lewis2020retrieval,
  title={Retrieval-augmented generation for knowledge-intensive nlp tasks},
  author={Lewis, Patrick and Perez, Ethan and Piktus, Aleksandra and Petroni, Fabio and Karpukhin, Vladimir and Goyal, Naman and K{\"u}ttler, Heinrich and Lewis, Mike and Yih, Wen-tau and Rockt{\"a}schel, Tim and others},
  journal={Advances in neural information processing systems},
  volume={33},
  pages={9459--9474},
  year={2020}
}

@article{nakano2021webgpt,
  title={Webgpt: Browser-assisted question-answering with human feedback, 2022},
  author={Nakano, Reiichiro and Hilton, Jacob and Balaji, Suchir and Wu, Jeff and Ouyang, Long and Kim, Christina and Hesse, Christopher and Jain, Shantanu and Kosaraju, Vineet and Saunders, William and others},
  journal={URL https://arxiv. org/abs/2112.09332},
  volume={35},
  year={2022}
}

@inproceedings{qin2023webcpm,
  title={Webcpm: Interactive web search for chinese long-form question answering},
  author={Qin, Yujia and Cai, Zihan and Jin, Dian and Yan, Lan and Liang, Shihao and Zhu, Kunlun and Lin, Yankai and Han, Xu and Ding, Ning and Wang, Huadong and others},
  booktitle={Proceedings of the 61st Annual Meeting of the Association for Computational Linguistics (Volume 1: Long Papers)},
  pages={8968--8988},
  year={2023}
}

@inproceedings{liu2023webglm,
  title={WebGLM: towards an efficient web-enhanced question answering system with human preferences},
  author={Liu, Xiao and Lai, Hanyu and Yu, Hao and Xu, Yifan and Zeng, Aohan and Du, Zhengxiao and Zhang, Peng and Dong, Yuxiao and Tang, Jie},
  booktitle={Proceedings of the 29th ACM SIGKDD conference on knowledge discovery and data mining},
  pages={4549--4560},
  year={2023}
}

@inproceedings{trivedi2023interleaving,
  title={Interleaving retrieval with chain-of-thought reasoning for knowledge-intensive multi-step questions},
  author={Trivedi, Harsh and Balasubramanian, Niranjan and Khot, Tushar and Sabharwal, Ashish},
  booktitle={Proceedings of the 61st annual meeting of the association for computational linguistics (volume 1: long papers)},
  pages={10014--10037},
  year={2023}
}

@inproceedings{asai2024selfrag,
  title={Self-rag: Learning to retrieve, generate, and critique through self-reflection},
  author={Asai, Akari and Wu, Zeqiu and Wang, Yizhong and Sil, Avi and Hajishirzi, Hannaneh},
  booktitle={International conference on learning representations},
  volume={2024},
  pages={9112--9141},
  year={2024}
}

@inproceedings{shao2024assisting,
  title={Assisting in writing wikipedia-like articles from scratch with large language models},
  author={Shao, Yijia and Jiang, Yucheng and Kanell, Theodore and Xu, Peter and Khattab, Omar and Lam, Monica},
  booktitle={Proceedings of the 2024 Conference of the North American Chapter of the Association for Computational Linguistics: Human Language Technologies (Volume 1: Long Papers)},
  pages={6252--6278},
  year={2024}
}

@inproceedings{zheng2025deepresearcher,
  title={Deepresearcher: Scaling deep research via reinforcement learning in real-world environments},
  author={Zheng, Yuxiang and Fu, Dayuan and Hu, Xiangkun and Cai, Xiaojie and Ye, Lyumanshan and Lu, Pengrui and Liu, Pengfei},
  booktitle={Proceedings of the 2025 Conference on Empirical Methods in Natural Language Processing},
  pages={414--431},
  year={2025}
}

@article{li2025webthinker,
  title={Webthinker: Empowering large reasoning models with deep research capability},
  author={Li, Xiaoxi and Jin, Jiajie and Dong, Guanting and Qian, Hongjin and Wu, Yongkang and Wen, Ji-Rong and Zhu, Yutao and Dou, Zhicheng},
  journal={Advances in Neural Information Processing Systems},
  volume={38},
  pages={120091--120131},
  year={2026}
}

@inproceedings{guu2020realm,
  title={Retrieval augmented language model pre-training},
  author={Guu, Kelvin and Lee, Kenton and Tung, Zora and Pasupat, Panupong and Chang, Mingwei},
  booktitle={International conference on machine learning},
  pages={3929--3938},
  year={2020},
  organization={PMLR}
}

@inproceedings{izacard2021leveraging,
  title={Leveraging passage retrieval with generative models for open domain question answering},
  author={Izacard, Gautier and Grave, Edouard},
  booktitle={Proceedings of the 16th conference of the european chapter of the association for computational linguistics: main volume},
  pages={874--880},
  year={2021}
}

@inproceedings{komeili2022internet,
  title={Internet-augmented dialogue generation},
  author={Komeili, Mojtaba and Shuster, Kurt and Weston, Jason},
  booktitle={Proceedings of the 60th annual meeting of the Association for Computational Linguistics (Volume 1: Long papers)},
  pages={8460--8478},
  year={2022}
}

@inproceedings{chen2020hybridqa,
  title={HybridQA: A dataset of multi-hop question answering over tabular and textual data},
  author={Chen, Wenhu and Zha, Hanwen and Chen, Zhiyu and Xiong, Wenhan and Wang, Hong and Wang, William Yang},
  booktitle={Findings of the Association for Computational Linguistics: EMNLP 2020},
  pages={1026--1036},
  year={2020}
}

@inproceedings{java2026characterizing,
  title={Characterizing deep research: A benchmark and formal definition},
  author={Java, Abhinav and Khandelwal, Ashmit and Midigeshi, Sukruta and Halfaker, Aaron and Deshpande, Amit Jayant and Goyal, Navin and Gupta, Ankur and Natarajan, Nagarajan and Sharma, Amit},
  booktitle={International Conference on Learning Representations},
  volume={2026},
  pages={10315--10349},
  year={2026}
}

@article{kwiatkowski2019naturalquestions,
  title={Natural questions: a benchmark for question answering research},
  author={Kwiatkowski, Tom and Palomaki, Jennimaria and Redfield, Olivia and Collins, Michael and Parikh, Ankur and Alberti, Chris and Epstein, Danielle and Polosukhin, Illia and Devlin, Jacob and Lee, Kenton and others},
  journal={Transactions of the Association for Computational Linguistics},
  volume={7},
  pages={453--466},
  year={2019},
  publisher={MIT Press One Rogers Street, Cambridge, MA 02142-1209, USA journals-info~…}
}

@inproceedings{joshi2017triviaqa,
  title={Triviaqa: A large scale distantly supervised challenge dataset for reading comprehension},
  author={Joshi, Mandar and Choi, Eunsol and Weld, Daniel S and Zettlemoyer, Luke},
  booktitle={Proceedings of the 55th Annual Meeting of the Association for Computational Linguistics (Volume 1: Long Papers)},
  pages={1601--1611},
  year={2017}
}

@inproceedings{mialon2024gaia,
  title={Gaia: a benchmark for general ai assistants},
  author={Mialon, Gr{\'e}goire and Fourrier, Cl{\'e}mentine and Wolf, Thomas and LeCun, Yann and Scialom, Thomas},
  booktitle={International Conference on Learning Representations},
  volume={2024},
  pages={9025--9049},
  year={2024}
}

@inproceedings{yoran2024assistantbench,
  title={Assistantbench: Can web agents solve realistic and time-consuming tasks?},
  author={Yoran, Ori and Amouyal, Samuel Joseph and Malaviya, Chaitanya and Bogin, Ben and Press, Ofir and Berant, Jonathan},
  booktitle={Proceedings of the 2024 Conference on Empirical Methods in Natural Language Processing},
  pages={8938--8968},
  year={2024}
}

@article{chen2025medbrowsecomp,
  title={Medbrowsecomp: Benchmarking medical deep research and computer use},
  author={Chen, Shan and Moreira, Pedro and Xiao, Yuxin and Schmidgall, Sam and Warner, Jeremy and Aerts, Hugo and Hartvigsen, Thomas and Gallifant, Jack and Bitterman, Danielle S},
  journal={arXiv preprint arXiv:2505.14963},
  year={2025}
}

@article{chen2021ottqa,
  title={Open question answering over tables and text},
  author={Chen, Wenhu and Chang, Ming-Wei and Schlinger, Eva and Wang, William and Cohen, William W},
  journal={arXiv preprint arXiv:2010.10439},
  year={2020}
}

@inproceedings{zhu2021tatqa,
  title={TAT-QA: A question answering benchmark on a hybrid of tabular and textual content in finance},
  author={Zhu, Fengbin and Lei, Wenqiang and Huang, Youcheng and Wang, Chao and Zhang, Shuo and Lv, Jiancheng and Feng, Fuli and Chua, Tat-Seng},
  booktitle={Proceedings of the 59th annual meeting of the Association for Computational Linguistics and the 11th international joint conference on natural language processing (volume 1: long papers)},
  pages={3277--3287},
  year={2021}
}

@article{zhong2017seq2sql,
  title={Seq2sql: Generating structured queries from natural language using reinforcement learning},
  author={Zhong, Victor and Xiong, Caiming and Socher, Richard},
  journal={arXiv preprint arXiv:1709.00103},
  year={2017}
}

@inproceedings{yu2019sparc,
  title={Sparc: Cross-domain semantic parsing in context},
  author={Yu, Tao and Zhang, Rui and Yasunaga, Michihiro and Tan, Yi Chern and Lin, Xi Victoria and Li, Suyi and Er, Heyang and Li, Irene and Pang, Bo and Chen, Tao and others},
  booktitle={Proceedings of the 57th annual meeting of the association for computational linguistics},
  pages={4511--4523},
  year={2019}
}

@inproceedings{yu2019cosql,
  title={Cosql: A conversational text-to-sql challenge towards cross-domain natural language interfaces to databases},
  author={Yu, Tao and Zhang, Rui and Er, Heyang and Li, Suyi and Xue, Eric and Pang, Bo and Lin, Xi Victoria and Tan, Yi Chern and Shi, Tianze and Li, Zihan and others},
  booktitle={Proceedings of the 2019 conference on empirical methods in natural language processing and the 9th international joint conference on natural language processing (EMNLP-IJCNLP)},
  pages={1962--1979},
  year={2019}
}

@article{li2025swe,
  title={Swe-sql: Illuminating llm pathways to solve user sql issues in real-world applications},
  author={Li, Jinyang and Li, Xiaolong and Qu, Ge and Jacobsson, Per and Qin, Bowen and Hui, Binyuan and Si, Shuzheng and Huo, Nan and Xu, Xiaohan and Zhang, Yue and others},
  journal={Advances in Neural Information Processing Systems},
  volume={38},
  pages={97085--97120},
  year={2026}
}

@inproceedings{huo2026bird,
  title={BIRD-INTERACT: Re-imagining Text-to-SQL Evaluation via Lens of Dynamic Interactions},
  author={Huo, Nan and Xu, Xiaohan and Li, Jinyang and Jacobsson, Per and Lin, Shipei and Qin, Bowen and Hui, Binyuan and Li, Xiaolong and Qu, Ge and Si, Shuzheng and others},
  booktitle={International Conference on Learning Representations},
  volume={2026},
  pages={44285--44333},
  year={2026}
}

@inproceedings{jiang2023active,
  title={Active retrieval augmented generation},
  author={Jiang, Zhengbao and Xu, Frank F and Gao, Luyu and Sun, Zhiqing and Liu, Qian and Dwivedi-Yu, Jane and Yang, Yiming and Callan, Jamie and Neubig, Graham},
  booktitle={Proceedings of the 2023 conference on empirical methods in natural language processing},
  pages={7969--7992},
  year={2023}
}

@inproceedings{jeong2024adaptive,
  title={Adaptive-rag: Learning to adapt retrieval-augmented large language models through question complexity},
  author={Jeong, Soyeong and Baek, Jinheon and Cho, Sukmin and Hwang, Sung Ju and Park, Jong C},
  booktitle={Proceedings of the 2024 conference of the north american chapter of the association for computational linguistics: Human language technologies (volume 1: Long papers)},
  pages={7036--7050},
  year={2024}
}

@inproceedings{li2025searcho1,
  title={Search-o1: Agentic search-enhanced large reasoning models},
  author={Li, Xiaoxi and Dong, Guanting and Jin, Jiajie and Zhang, Yuyao and Zhou, Yujia and Zhu, Yutao and Zhang, Peitian and Dou, Zhicheng},
  booktitle={Proceedings of the 2025 Conference on Empirical Methods in Natural Language Processing},
  pages={5420--5438},
  year={2025}
}

@inproceedings{wu2025webwalker,
  title={Webwalker: Benchmarking llms in web traversal},
  author={Wu, Jialong and Yin, Wenbiao and Jiang, Yong and Wang, Zhenglin and Xi, Zekun and Fang, Runnan and Zhang, Linhai and He, Yulan and Zhou, Deyu and Xie, Pengjun and others},
  booktitle={Proceedings of the 63rd Annual Meeting of the Association for Computational Linguistics (Volume 1: Long Papers)},
  pages={10290--10305},
  year={2025}
}

@inproceedings{song2026demystifying,
  title={Demystifying deep search: a holistic evaluation with hint-free multi-hop questions and factorised metrics},
  author={Song, Maojia and Renhang, Liu and Wang, Xinyu and Jiang, Yong and Xie, Pengjun and Huang, Fei and Poria, Soujanya and Zhou, Jingren},
  booktitle={International Conference on Learning Representations},
  volume={2026},
  pages={95487--95512},
  year={2026}
}

@inproceedings{paul2026deepsynth,
  title={A Benchmark for Deep Information Synthesis},
  author={Paul, Debjit and Murphy, Daniel and Gritta, Milan and Cardenas Acosta, Ronald and Prokhorov, Victor and Bolliger, Lena Sophia and Toker, Aysim and Miles, Roy and Oncescu, Andreea-Maria and Sivakumar, Jasivan and others},
  booktitle={International Conference on Learning Representations},
  volume={2026},
  pages={65663--65684},
  year={2026}
}

@inproceedings{christmann2024compmix,
  title={Compmix: A benchmark for heterogeneous question answering},
  author={Christmann, Philipp and Saha Roy, Rishiraj and Weikum, Gerhard},
  booktitle={Companion Proceedings of the ACM Web Conference 2024},
  pages={1091--1094},
  year={2024}
}

@inproceedings{li2021dual,
  title={Dual reader-parser on hybrid textual and tabular evidence for open domain question answering},
  author={Li, Alexander Hanbo and Ng, Patrick and Xu, Peng and Zhu, Henghui and Wang, Zhiguo and Xiang, Bing},
  booktitle={Proceedings of the 59th Annual Meeting of the Association for Computational Linguistics and the 11th International Joint Conference on Natural Language Processing (Volume 1: Long Papers)},
  pages={4078--4088},
  year={2021}
}

@article{biswal2025text2sql,
  title={Text2sql is not enough: Unifying ai and databases with tag},
  author={Biswal, Asim and Patel, Liana and Jha, Siddarth and Kamsetty, Amog and Liu, Shu and Gonzalez, Joseph E and Guestrin, Carlos and Zaharia, Matei},
  journal={arXiv preprint arXiv:2408.14717},
  year={2024}
}

@article{zhao2025hybrid,
  title={Hybrid querying over relational databases and large language models},
  author={Zhao, Fuheng and Agrawal, Divyakant and Abbadi, Amr El},
  journal={arXiv preprint arXiv:2408.00884},
  year={2024}
}

@article{ning2026mcsearch,
  title={MC-Search: Evaluating and Enhancing Multimodal Agentic Search with Structured Long Reasoning Chains},
  author={Ning, Xuying and Fu, Dongqi and Wei, Tianxin and Ai, Mengting and Zou, Jiaru and Li, Ting-Wei and Tong, Hanghang and Zhu, Yada and Hamann, Hendrik and He, Jingrui},
  journal={arXiv preprint arXiv:2603.00873},
  year={2026}
}

@inproceedings{choubey2025benchmarking,
  title={Benchmarking deep search over heterogeneous enterprise data},
  author={Choubey, Prafulla Kumar and Peng, Xiangyu and Bhagavath, Shilpa and Huang, Kung-Hsiang and Xiong, Caiming and Wu, Chien-Sheng},
  booktitle={Proceedings of the 2025 Conference on Empirical Methods in Natural Language Processing: Industry Track},
  pages={501--517},
  year={2025}
}

@inproceedings{abaskohi2026drbench,
  title={Drbench: A realistic benchmark for enterprise deep research},
  author={Abaskohi, Amirhossein and Chen, Tianyi and Mu{\~n}oz-M{\'a}rmol, Miguel and Fox, Curtis and Ramesh, Amrutha Varshini and Marcotte, {\'E}tienne and L{\`u}, Xing Han and Chapados, Nicolas and Gella, Spandana and Pal, Christopher and others},
  booktitle={International Conference on Learning Representations},
  volume={2026},
  pages={6727--6795},
  year={2026}
}

@inproceedings{ferreira2026deepresearchretail,
  title={DeepResearch Retail: Benchmarking Tool-Augmented Deep Research in the E-Commerce Domain},
  author={Ferreira, Rafael and Di Palo, Flavio and Lu, Huilin and Jain, Ayush and Aduri, Harsha},
  booktitle={Proceedings of the 64th Annual Meeting of the Association for Computational Linguistics (ACL 2026)},
  pages={386--409},
  year={2026}
}

@article{zhou2025browsecompzh,
  title={Browsecomp-zh: Benchmarking web browsing ability of large language models in chinese},
  author={Zhou, Peilin and Leon, Bruce and Ying, Xiang and Zhang, Can and Shao, Yifan and Ye, Qichen and Chong, Dading and Jin, Zhiling and Xie, Chenxuan and Cao, Meng and others},
  journal={arXiv preprint arXiv:2504.19314},
  year={2025}
}

@article{li2025mmbrowsecomp,
  title={Mm-browsecomp: A comprehensive benchmark for multimodal browsing agents},
  author={Li, Shilong and Bu, Xingyuan and Wang, Wenjie and Liu, Jiaheng and Dong, Jun and He, Haoyang and Lu, Hao and Zhang, Haozhe and Jing, Chenchen and Li, Zhen and others},
  journal={arXiv preprint arXiv:2508.13186},
  year={2025}
}

@article{bosse2025deepresearchbench,
  title={Deep research bench: Evaluating ai web research agents},
  author={Bosse, Nikos I and Evans, Jon and Gambee, Robert G and Hnyk, Daniel and M{\"u}hlbacher, Peter and Phillips, Lawrence and Schwarz, Dan and Wildman, Jack and others},
  journal={arXiv preprint arXiv:2506.06287},
  year={2025}
}

@article{liu2026datastorm,
  title={DataSTORM: Deep Research on Large-Scale Databases using Exploratory Data Analysis and Data Storytelling},
  author={Liu, Shicheng and Jiang, Yucheng and Farook, Sajid and Sanchez, Camila Nicollier and Pena, David Fernando Castro and Lam, Monica S},
  journal={arXiv preprint arXiv:2604.06474},
  year={2026}
}

@misc{openai2025gpt5,
  author = {{OpenAI}},
  title  = {{GPT-5 System Card}},
  year   = {2025},
  url    = {https://openai.com/index/gpt-5-system-card/}
}

@misc{anthropic2026sonnet46,
  author = {{Anthropic}},
  title  = {Introducing {Claude Sonnet 4.6}},
  year   = {2026},
  url    = {https://www.anthropic.com/news/claude-sonnet-4-6}
}

\newpage
\appendix
\section{Detailed Hybrid Examples}
\label{app:examples}

We include one complete example for each hybrid type.
For each example we show the combined question, the search and SQL subcomponents with gold outputs, and the final answer.

\subsection{\texttt{S2SQL}: \texttt{s2sql\_000}}
The agent must first resolve multi-clue web evidence to the pivot \emph{Monaco}, then run SQL that ranks Monaco-related data flows by reliability.

\paragraph{Combined question.}
\begin{quote}
\small
Could you identify the top 5 data flows by reliability score, and show how each compares to the average reliability score within its channel protocol, for flows involving a place that:\\
- is where a French singer-actor born in Saint-Cyr-l'École died\\
- was the setting for a discussion involving a London-born racing driver and presenter at the 2012 Historic Grand Prix there\\
- comes up in travel comparisons with a French commune in Bouches-du-Rhône, which is described as having lower average daily costs\\
- shares its name with the surname of a Philadelphia-born television actor and model\\
- is also echoed by the name of a Devon village and civil parish mentioned alongside that actor and model\\
~\\
For each of the top 5 flows, return the flow tag, the channel protocol, the flow's reliability score, and the average reliability score across all flows in that same channel protocol. Use the in cross\_db database records to answer this.
\end{quote}

\paragraph{Search subquestion.}
\begin{quote}
\small
Could you provide me with the name of the place that:\\
- is where a French singer-actor born in Saint-Cyr-l'École died\\
- was the setting for a discussion involving a London-born racing driver and presenter at the 2012 Historic Grand Prix there\\
- comes up in travel comparisons with a French commune in Bouches-du-Rhône, which is described as having lower average daily costs\\
- shares its name with the surname of a Philadelphia-born television actor and model\\
- is also echoed by the name of a Devon village and civil parish mentioned alongside that actor and model
\end{quote}

\noindent Gold search answer: \emph{Monaco}.

\paragraph{SQL subquestion.}
\begin{quote}
\small
For data flows involving Monaco, what are the top 5 flows by their reliability score, and how does each compare to the average reliability score within its channel protocol?
\end{quote}

\begin{lstlisting}
SELECT flowtag, chanproto,
  ROUND((successpct / (errtally + 1.0))
    * (1.0 - rtrytally / (errtally + 1.0)), 4)
    AS dfrs,
  ROUND(AVG((successpct / (errtally + 1.0))
    * (1.0 - rtrytally / (errtally + 1.0)))
    OVER (PARTITION BY chanproto), 4)
    AS proto_avg_dfrs
FROM dataflow
WHERE destnation = 'Monaco'
   OR orignation = 'Monaco'
ORDER BY dfrs DESC
LIMIT 5
\end{lstlisting}
\noindent Gold SQL / final answer:

{\footnotesize
\begin{center}
\begin{tabular}{llrr}
\toprule
\texttt{flowtag} & \texttt{chanproto} & \texttt{dfrs} & \texttt{proto\_avg\_dfrs} \\
\midrule
DF7316 & Private Network & 2.0498 & -10.0122 \\
DF3144 & Private Network & 0.9787 & -10.0122 \\
DF1931 & Blockchain & 0.7845 & 0.7845 \\
DF4213 & HTTPS & 0.7491 & 0.7491 \\
DF5144 & Private Network & 0.5418 & -10.0122 \\
\bottomrule
\end{tabular}
\end{center}
}

\subsection{\texttt{SQL2S}: \texttt{sql2s\_105}}
SQL first identifies the pivot country \emph{Bhutan} under session constraints.
Search then recovers the ethnic group \emph{Lhop people} under Bhutan-specific clues.

\paragraph{Combined question.}
\begin{quote}
\small
In the news database, consider countries that have exactly 4 sessions, all of which fall into the average performance segment, and whose mean engagement score sits moderately between 0.4 and 0.5. Among these, the country that leads in average session duration is the focus of this question. Please find the name of the ethnic group which lives in that country, also in a human settlement there and near a border city in a district of that same country. It is referred to in a modern language used in that country and in India, and its dress resembles that of a South Asian community spanning that country, Nepal, and India. What is the name of this group?
\end{quote}

\paragraph{SQL subquestion.}
\begin{quote}
\small
Among countries with exactly 4 sessions that all fall into the average performance segment and have a moderate mean engagement score between 0.4 and 0.5, which country leads in average session duration?
\end{quote}

\begin{lstlisting}
WITH adjusted AS (
  SELECT seshkey, geoctry, bncrate, ctrval,
    engscore, seshdur,
    bncrate * (1.0 - ctrval / 100.0)
      AS adjusted_bounce_rate
  FROM sessions
),
ranked AS (
  SELECT seshkey, geoctry,
    adjusted_bounce_rate, ctrval,
    engscore, seshdur,
    PERCENT_RANK() OVER (
      ORDER BY adjusted_bounce_rate)
      AS bounce_percentile,
    PERCENT_RANK() OVER (
      ORDER BY ctrval DESC)
      AS ctr_percentile
  FROM adjusted
),
segmented AS (
  SELECT seshkey, geoctry,
    engscore, seshdur,
    CASE
      WHEN bounce_percentile < 0.25
        AND ctr_percentile < 0.25
        THEN 'High Bounce, Low CTR'
      WHEN bounce_percentile < 0.25
        AND ctr_percentile >= 0.75
        THEN 'High Bounce, High CTR'
      WHEN bounce_percentile >= 0.75
        AND ctr_percentile < 0.25
        THEN 'Low Bounce, Low CTR'
      WHEN bounce_percentile >= 0.75
        AND ctr_percentile >= 0.75
        THEN 'Low Bounce, High CTR'
      ELSE 'Average Performance'
    END AS perf_segment
  FROM ranked
),
country_agg AS (
  SELECT geoctry,
    COUNT(*) AS total_sessions,
    SUM(CASE perf_segment
      WHEN 'Average Performance'
      THEN 1 ELSE 0 END)
      AS avg_perf_count,
    ROUND(AVG(seshdur), 2)
      AS avg_session_duration,
    ROUND(AVG(engscore), 4)
      AS avg_engagement
  FROM segmented
  GROUP BY geoctry
  HAVING total_sessions = 4
     AND avg_perf_count = 4
     AND avg_engagement
         BETWEEN 0.4 AND 0.5
)
SELECT geoctry
FROM country_agg
ORDER BY avg_session_duration DESC
LIMIT 1
\end{lstlisting}
\noindent Gold SQL answer: \emph{Bhutan}.

\paragraph{Search subquestion.}
\begin{quote}
\small
Please find the name of the ethnic group which lives in Bhutan, also in a Bhutanese human settlement and near a Bhutanese border city in a district. It is referred to in a modern language used in Bhutan and India, and its dress resembles that of a South Asian community spanning Bhutan, Nepal, and India. What is the name of this group?
\end{quote}

\noindent Gold search / final answer: \emph{Lhop people}.

\subsection{\texttt{Parallel}: \texttt{parallel\_048}}
Search and SQL independently produce candidate country sets.
The final answer is their intersection, which isolates \emph{Finland}.

\paragraph{Combined question.}
\begin{quote}
\small
Could you provide me with the name of the country who:\\
- contains both a neighborhood rapid-transit stop in a capital city and an Olympic football venue in a capital district, with those two places being in the same capital\\
- is the place people move to when a state organ headquartered in an office centre handles residence registration and startup permit applications\\
- has as its official investment promotion body a nonprofit government agency\\
- is the citizenship of a person in ski jumping and Nordic combined\\
- ranks among the top 10 countries by their share of high-CTR performance segment sessions in news database, considering only countries with at least 5 sessions total and at least 2 sessions in both the high-CTR segments and the average performance segment
\end{quote}

\paragraph{Search subquestion.}
\begin{quote}
\small
Could you provide me with the name of the country who:\\
- contains both a neighborhood rapid-transit stop in a capital city and an Olympic football venue in a capital district, with those two places being in the same capital\\
- is the place people move to when a state organ headquartered in an office centre handles residence registration and startup permit applications\\
- has as its official investment promotion body a nonprofit government agency\\
- is the citizenship of a person in ski jumping and Nordic combined
\end{quote}

\noindent Gold search set: $\{$Finland, Estonia$\}$.

\paragraph{SQL subquestion.}
\begin{quote}
\small
Which are the top 10 countries by their share of high-CTR performance segment sessions, considering only countries with at least 5 sessions and at least 2 sessions in both the high-CTR segments and the average performance segment?
\end{quote}

\begin{lstlisting}
WITH adjusted AS (
  SELECT seshkey, geoctry,
    bncrate * (1 - ctrval/100.0)
      AS adjusted_bounce_rate,
    ctrval,
    PERCENT_RANK() OVER (
      ORDER BY bncrate * (1 - ctrval/100.0))
      AS bounce_percentile,
    PERCENT_RANK() OVER (
      ORDER BY ctrval DESC)
      AS ctr_percentile
  FROM sessions
),
segmented AS (
  SELECT seshkey, geoctry,
    CASE
      WHEN bounce_percentile < 0.25
        AND ctr_percentile < 0.25
        THEN 'High Bounce, Low CTR'
      WHEN bounce_percentile < 0.25
        AND ctr_percentile >= 0.75
        THEN 'High Bounce, High CTR'
      WHEN bounce_percentile >= 0.75
        AND ctr_percentile < 0.25
        THEN 'Low Bounce, Low CTR'
      WHEN bounce_percentile >= 0.75
        AND ctr_percentile >= 0.75
        THEN 'Low Bounce, High CTR'
      ELSE 'Average Performance'
    END AS perf_segment
  FROM adjusted
),
country_stats AS (
  SELECT geoctry,
    COUNT(*) AS total_sessions,
    SUM(CASE WHEN perf_segment IN (
      'High Bounce, High CTR',
      'Low Bounce, High CTR')
      THEN 1 ELSE 0 END)
      AS high_ctr_sessions,
    SUM(CASE WHEN perf_segment =
      'Average Performance'
      THEN 1 ELSE 0 END)
      AS avg_sessions
  FROM segmented
  GROUP BY geoctry
  HAVING COUNT(*) >= 5
)
SELECT geoctry
FROM country_stats
WHERE high_ctr_sessions >= 2
  AND avg_sessions >= 2
ORDER BY CAST(high_ctr_sessions AS REAL)
  / total_sessions DESC
LIMIT 10
\end{lstlisting}
\noindent Gold SQL set (top countries): $\{$Northern Mariana Islands,
Finland, Cambodia, Faroe Islands, Jordan, Somalia,
Switzerland, Italy, New Caledonia, Puerto Rico$\}$.

\noindent Final answer: \emph{Finland}, the intersection of the two sets.

\section{Filtering Details}
\label{app:filtering}

Before cross-modal composition, we filter the search and SQL components independently.

\subsection{Search Component Filtering}

Each search question passes five stages.

\paragraph{Closed-book prefilter.}
We remove questions that can already be answered by a closed-book solver, reducing reliance on parametric memory.

\paragraph{Search quality.}
The question must resist naive whole-question web search and BM25 retrieval.
At the same time, every reasoning step must remain recoverable through a targeted query that excludes the gold answer and its aliases.
Evidence presence is determined through substring matching or an LLM-based snippet judge.

\paragraph{Clue integrity.}
We verify that rewritten clues neither reveal nor contradict the source entities.
The answer must also be recoverable from the evidence subgraph alone.

\paragraph{Candidate inference.}
We collect candidate answers from multiple base models and decoding temperatures.

\paragraph{Answer cardinality.}
We deduplicate the candidate answers and verify each candidate against all question constraints without exposing the gold answer.

The retained questions are then assigned to hybrid patterns.
Questions that contain the seed in the question support \texttt{SQL2S}.
When the seed appears in the answer set, a single verified answer supports \texttt{S2SQL}, while multiple verified answers support \texttt{Parallel}.
Questions that fail any stage or produce candidates inconsistent with the evidence constraints are discarded.

\subsection{SQL Component Filtering}

Each SQL question-query pair must execute successfully and return a non-empty result.
We then verify that the seed plays the role required by the target hybrid pattern.

For \texttt{S2SQL}, the seed must appear in the question as a constraint.
For \texttt{SQL2S}, the query must return a single row containing the seed.
For \texttt{Parallel}, the query must return multiple rows, including one that contains the seed.

Pairs that fail execution or role checks are removed.
Human reviewers then retain the higher-quality and more challenging SQL components and discard ambiguous, shallow, or low-value candidates.

\section{Human Review Details}
\label{app:human-review}

Human review was conducted by four doctoral researchers who are also authors of this paper.
The review covered SQL pattern validation, final question quality control, and balanced hard-subset selection.

\paragraph{SQL pattern validation.}
The reviewers examined the catalog for structural validity, diversity, and compatibility with the source databases.
Redundant, invalid, or overly narrow patterns were removed or revised.

\paragraph{Final question quality control.}
The reviewers examined the retained questions after automatic filtering and cross-modal composition.
They checked whether each question was clear, whether the gold answer was correct and unique, whether the search and SQL components were mutually consistent, and whether the shared entity played the intended role in the target reasoning pattern.
They also checked for unintended answer leakage and removed or revised questions that failed these criteria.

\paragraph{Hard-subset selection.}
The reviewers used intrinsic task complexity and pilot agent evaluation results as complementary signals.
They considered the complexity of the search and SQL components, the number and specificity of constraints transferred between tools, and recurring failure patterns observed during pilot evaluation.
Pilot results were used to help identify challenging questions rather than as the sole selection criterion.
The selected questions were then manually verified, resulting in a balanced subset of $40$ questions from each reasoning pattern.

\paragraph{Review process.}
Review decisions were discussed among the four reviewers, and disagreements were resolved through consensus.

\section{Experimental Details}
\label{app:exp-details}

\paragraph{Agent frameworks.}
In smolagents~\cite{smolagents}, a single tool-calling agent solves each task within a shared context.
In MiroFlow~\cite{su2026miroflow}, a main agent delegates subtasks to specialized workers through a hierarchical agent graph.
Both frameworks receive the same task requirements, adapted to their respective prompting formats, and access to the same read-only SQL and knowledge-catalog tools.
Detailed tool specifications are provided in \apref{app:tools}.

\paragraph{Retrieval backends.}
Real Web searches the live web.
Corpus Search retrieves from a fixed index of Wikipedia and FineWeb-10BT~\cite{penedo2024fineweb}.
The fixed corpus provides a controlled retrieval setting aligned with the evidence sources used during benchmark construction.

\paragraph{Models and serving.}
Claude-Sonnet-4.6 and GPT-5 are accessed through their hosted APIs with medium reasoning effort.
For open-weight models, we evaluate Qwen3.5~\cite{qwen3.5} and GLM-5.2~\cite{zeng2026glm}.
The Qwen3.5 family includes the 397B-A17B and 122B-A10B MoE models, together with the 27B, 9B, and 4B dense models.
We serve the Qwen models in FP8 with vLLM~\cite{kwon2023efficient} on NVIDIA H200 and B200 GPUs.
We additionally evaluate GLM-5.2-FP4.

\paragraph{Inference settings.}
For both open-weight model families, we use the decoding hyperparameters recommended by the model providers. Specifically, Qwen3.5 uses temperature $0.6$, $\mathrm{top}\text{-}p{=}0.95$, and $\mathrm{top}\text{-}k{=}20$, with thinking enabled.
GLM-5.2 uses temperature $1.0$, $\mathrm{top}\text{-}p{=}0.95$, and its maximum reasoning-effort setting.
All agents are limited to at most $300$ tool-calling turns and a $3600$-second wall-clock budget per task.

\paragraph{Evaluation scope.}
Open-weight models are evaluated on the full set of $380$ tasks under both agent frameworks.
Claude-Sonnet-4.6 and GPT-5 are evaluated on the balanced $120$-example hard subset with smolagents.
The hard subset contains $40$ tasks from each hybrid type.

\paragraph{Task-specific evaluation.}
For \texttt{S2SQL}, we extract the SQL query produced by the agent and execute both the predicted and gold queries on the corresponding LiveSQLBench~\cite{livesqlbench2025} SQLite database.
The resulting tables are compared after column alignment.

For \texttt{SQL2S} and \texttt{Parallel}, we use a BrowseComp-style LLM judge~\cite{wei2025browsecomp}.
The judge compares the predicted free-text answer with the gold answer and uses Qwen3-30B-A3B-Instruct-2507 served through vLLM~\cite{kwon2023efficient}.

\paragraph{Repeated trials.}
Each task is evaluated over eight independent trials.
Pass@8~\cite{chen2021evaluating} reports whether a task is solved in at least one trial.
Avg@8 reports the average success rate across the eight trials.

\section{Evaluation Tool Interface}
\label{app:tools}

Both \emph{smolagents} and \emph{MiroFlow} are given the same hybrid tool contract.
Tools fall into three groups: structured SQL / knowledge lookup (always available), and one of two mutually exclusive search backends (Real Web or Corpus Search).
Below we list each tool name, arguments, return format, and underlying service.

\subsection{Structured Tools}

\paragraph{\texttt{run\_sql}.}
Execute a read-only query against the task's LiveSQLBench~\cite{livesqlbench2025} SQLite database.
\textbf{Inputs:} \texttt{sql} (string, required) is the query, and \texttt{description} (string, optional) is a one-line intent summary prepended to the result.
Only \texttt{SELECT} / \texttt{WITH} / \texttt{PRAGMA} / \texttt{EXPLAIN} are allowed. Write statements (\texttt{INSERT}, \texttt{UPDATE}, \texttt{DELETE}, \texttt{DROP}, \texttt{ALTER}, \texttt{CREATE}, \texttt{ATTACH}, \texttt{DETACH}, \texttt{VACUUM}, \texttt{REPLACE}) are rejected.
\textbf{Output:} a text table (column header + rows), truncated at 500 rows by default, or an error string.
\textbf{Backend:} local SQLite in read-only URI mode.
No schema dump is injected into the prompt. Agents must discover tables and columns via exploratory queries (e.g., \texttt{sqlite\_master}, \texttt{PRAGMA table\_info}).

\paragraph{\texttt{get\_all\_external\_knowledge\_names}.}
List domain knowledge titles for the current database.
Takes no inputs and returns a JSON list of titles, backed by the per-database LiveSQLBench knowledge file \texttt{\{db\}\_kb.jsonl}.

\paragraph{\texttt{get\_knowledge\_definition}.}
Fetch one knowledge entry in full.
\textbf{Input:} \texttt{knowledge\_name} (string, required): entry title or id (numeric or database-qualified, e.g., \texttt{alien\_0}).
\textbf{Output:} a JSON object with fields \texttt{id}, \texttt{knowledge}, \texttt{description}, \texttt{definition}, \texttt{type}, and \texttt{children\_knowledge}, or a not-found message.

\paragraph{\texttt{get\_all\_knowledge\_definitions}.}
Return the full knowledge catalog for the database (no inputs).
\textbf{Output:} a JSON array of full entries. If the payload exceeds a context budget, a titles-only summary is returned instead.

\subsection{Real Web Backend}

Used when agents retrieve from the live public web.

\paragraph{\texttt{web\_search}.}
\textbf{Inputs:} \texttt{query} (string, required, $\le$400 characters / 50 words) and \texttt{count} (int, default 10, capped at 50).
\textbf{Output:} a numbered list of \texttt{\{title, url, description\}} hits, prefixed by the backend label.
\textbf{Backend:} Serper Google Search API (\url{https://serper.dev/}).

\paragraph{\texttt{web\_fetch}.}
\textbf{Input:} \texttt{url} (string, required).
\textbf{Output:} cleaned page text / markdown (truncated at $\sim$24k characters).
PDFs are auto-detected and routed to PDF extraction. Repeated fetches of the same URL within a run return a cache note.
\textbf{Backend:} primary path is Jina Reader (\url{https://jina.ai/}). Fallback is \texttt{requests} + MarkItDown HTML-to-markdown conversion.

\paragraph{\texttt{pdf\_read}.}
\textbf{Inputs:} \texttt{source} (string, required): URL or local path. \texttt{pages} (string, optional): page selection such as \texttt{1-5} or \texttt{3,7,10} (default: all pages).
\textbf{Output:} extracted PDF text (truncated at $\sim$32k characters).
\textbf{Backend:} \texttt{pypdf} for local bytes / downloaded PDFs, with Jina Reader as a URL fallback.

\paragraph{\texttt{read\_file}.}
\textbf{Inputs:} \texttt{file\_path} (string, required), \texttt{offset} (int, default 1, 1-based start line), and \texttt{limit} (int, default 2000).
\textbf{Output:} numbered local-file lines, or PDF text if the path points to a PDF.
\textbf{Backend:} local filesystem I/O (with the same PDF extractor as \texttt{pdf\_read}).

\subsection{Corpus Search Backend}

Used when agents retrieve from the closed corpus that grounds \OURDATA{} construction (Wikipedia + FineWeb-10BT~\cite{penedo2024fineweb}).
Retrieval is implemented as a two-stage Corpus Search call inside \texttt{web\_search}: a document-level search first retrieves candidate pages with full text, then a chunk-level search within those pages produces a short query-conditioned snippet (truncated to the first 100 characters).
\texttt{web\_search} returns \texttt{title}, \texttt{url}, \texttt{date}, and the chunk snippet to the agent, while caching the corresponding full document text in session.
\texttt{visit\_web} does not issue another Corpus Search query. It only reads that cache.
In this mode, live \texttt{web\_fetch} / \texttt{pdf\_read} / \texttt{read\_file} are disabled. The agent-facing search tool is still named \texttt{web\_search}.

\paragraph{\texttt{web\_search} (Corpus Search).}
\textbf{Inputs:} \texttt{query} (string, required) and \texttt{k} (int, default 10, capped at 20).
\textbf{Output:} up to $k$ documents with \texttt{title}, \texttt{url}, \texttt{date}, and a query-conditional chunk \texttt{snippet} (first 100 characters).
\textbf{Backend:} two-stage Corpus Search (document-level then chunk-level) over the Wikipedia/FineWeb index. Full documents are cached in-session for \texttt{visit\_web}.

\paragraph{\texttt{visit\_web}.}
\textbf{Input:} \texttt{url} (string, required): must come from a prior \texttt{web\_search} hit in the same run.
\textbf{Output:} a JSON object with \texttt{title}, \texttt{url}, \texttt{date}, and full indexed \texttt{text} (truncated at $\sim$40k characters).
\textbf{Backend:} session cache populated by \texttt{web\_search} (no additional Corpus Search call, no live crawl).

\section{Full-Set Proprietary Runs}
\label{app:full-api}

Due to API cost, Claude-Sonnet-4.6 and GPT-5 are evaluated with only two independent full-set trials under smolagents with Corpus Search.
\tref{tab:full-api} reports Pass@2 and Avg@2.
These numbers are not directly comparable to the Pass@8~/~Avg@8 protocol used in the main text.
On the full set, Claude-Sonnet-4.6 obtains higher overall Pass@2 and Avg@2 than GPT-5, reversing the subset ranking in \tref{tab:subset-smol}.
GPT-5 remains stronger on S2SQL, but S2SQL is only $58$ of $380$ full-set examples ($15.3\%$), versus one third of the balanced subset.
As a result, GPT-5's S2SQL advantage contributes less to the full-set overall score.

\begin{table*}[t]
\centering
\begin{tabular}{llcccc}
\toprule
\textbf{Model} & \textbf{Metric} & \textbf{All} & \textbf{S2SQL} & \textbf{SQL2S} & \textbf{Parallel} \\
\midrule
\multirow{2}{*}{Claude-Sonnet-4.6}
 & Pass@2 & \textbf{57.37} & 39.66 & \textbf{51.72} & \textbf{75.63} \\
 & Avg@2 & \textbf{49.34} & 29.31 & \textbf{44.09} & \textbf{68.07} \\
\midrule
\multirow{2}{*}{GPT-5}
 & Pass@2 & 54.47 & \textbf{46.55} & 44.33 & \textbf{75.63} \\
 & Avg@2 & 44.61 & \textbf{34.48} & 35.71 & 64.71 \\
\bottomrule
\end{tabular}
\caption{Full-set smolagents + Corpus Search results for proprietary models with two trials (Pass@2~/~Avg@2). Bold marks the best score in each column for Pass@2 and Avg@2.}
\label{tab:full-api}
\end{table*}

\section{Detailed Trajectory Analysis}
\label{app:trajectory-analysis}

We provide detailed interaction statistics and representative failure cases for Qwen3.5-397B-A17B on the balanced hard subset under Corpus Search.
The analysis uses completed trajectory logs from both smolagents and MiroFlow.

\subsection{Trajectory Cost}

\begin{table}[t]
\centering
\small
\setlength{\tabcolsep}{3.5pt}
\begin{tabular}{llrrrr}
\toprule
\textbf{Scaffold} & \textbf{Type}
& \textbf{\#LLM}
& \textbf{\#Tool}
& \textbf{In}
& \textbf{Out} \\
\midrule
\multirow{4}{*}{smolagents}
& All      & 52.5 & 78.1 & 2.78M & 34.9K \\
& S2SQL    & 40.0 & 66.8 & 2.02M & 23.2K \\
& SQL2S    & 65.9 & 84.3 & 3.69M & 35.8K \\
& Parallel & 54.9 & 84.8 & 2.85M & 46.0K \\
\midrule
\multirow{4}{*}{MiroFlow}
& All      & 120.1 & 210.0 & 2.06M & 46.1K \\
& S2SQL    & 105.1 & 157.8 & 1.61M & 39.0K \\
& SQL2S    & 119.5 & 222.1 & 2.09M & 45.9K \\
& Parallel & 139.1 & 263.0 & 2.59M & 54.9K \\
\bottomrule
\end{tabular}
\caption{
Trajectory statistics for Qwen3.5-397B-A17B on the balanced hard subset with Corpus Search.
\#LLM and \#Tool report mean calls per completed trajectory.
In and Out report mean cumulative input and output tokens.
}
\label{tab:traj-stats}
\end{table}

MiroFlow uses substantially more LLM and tool calls than smolagents because it delegates subtasks to multiple workers.
Its input-token count is nevertheless similar or lower, since each worker maintains a shorter local context.

Of the $960$ attempts per scaffold, $800$ smolagents runs and $564$ MiroFlow runs finish within the one-hour limit.
These counts correspond to timeout rates of $16.7\%$ and $41.3\%$, respectively.
MiroFlow therefore explores more broadly, but at a substantially higher completion cost.

\subsection{Representative Failure Cases}

\paragraph{Incomplete intersection.}
On \texttt{Parallel} tasks, agents may return a candidate supported by only one source.
For \texttt{parallel\_106}, smolagents repeatedly returns Alaska after finding it in the SQL result.
However, Alaska does not satisfy the web-search constraints.
MiroFlow correctly compares the two candidate sets and recovers Vermont as their intersection.

\paragraph{Search failure after a correct SQL result.}
On \texttt{SQL2S} tasks, the database step may recover the correct entity, but the following search may drift or exhaust its budget.
For \texttt{sql2s\_136}, the SQL query correctly identifies Malta.
However, smolagents fails to preserve this constraint during web search and returns Schnitzel instead of the intended answer, powidl.

\paragraph{Incorrect SQL after a correct search result.}
On \texttt{S2SQL} tasks, web search may identify the correct entity, while the final SQL query uses an incorrect aggregation or formula.
For \texttt{s2sql\_000}, both scaffolds correctly identify Monaco.
However, the computed reliability value does not match the gold SQL execution.

\subsection{Scaffold Comparison}

MiroFlow's additional worker attempts often help recover constraints that were missed earlier.
This is especially useful for directional tasks and candidate-set intersection.
However, the same exploration produces more calls and more timeouts.
This trade-off helps explain why MiroFlow improves Pass@8 while Avg@8 remains close to smolagents.

\section{Related Work}
\label{app:related-work}
\subsection{From Retrieval-Augmented Generation to Agentic Deep Research}

Retrieval-augmented generation (RAG) provides a language model with evidence retrieved from an external corpus to improve its performance on knowledge-intensive tasks \cite{guu2020realm,lewis2020retrieval,izacard2021leveraging}. As tasks become more complex, adaptive retrieval extends this paradigm by interleaving retrieval with reasoning across multiple steps \cite{trivedi2023interleaving,jiang2023active,jeong2024adaptive,asai2024selfrag}. However, fixed-corpus retrieval remains restrictive in real-world settings, as relevant evidence often falls outside predefined collections or evolves dynamically \cite{komeili2022internet,qin2023webcpm}. Web search mitigates this limitation by enabling open-world information seeking without assuming that all task-relevant evidence has been collected in advance \cite{nakano2021webgpt,liu2023webglm}. Deep research tasks push further along both dimensions: they demand processing a large number of information units, and locating or synthesizing those units requires non-trivial reasoning \cite{java2026characterizing}. Existing deep research systems rely primarily on web search for evidence acquisition \cite{shao2024assisting,zheng2025deepresearcher,li2025webthinker,li2025searcho1}, with limited exploration of database querying as part of the research process. \OURDATA{} targets this hybrid setting, focusing on tasks with substantial search and reasoning demands whose supporting evidence is distributed across both web pages and database records.

\subsection{Evaluating Deep Research Agents}

Evaluating models on tasks that require retrieving and reasoning over external information is a long-standing research problem. Early question-answering benchmarks focus on questions that can be answered by locating a small amount of textual evidence \cite{kwiatkowski2019naturalquestions,joshi2017triviaqa}. Multi-hop benchmarks increase reasoning complexity by requiring evidence to be combined across documents, but still operate over predefined text corpora \cite{yang2018hotpotqa,trivedi2022musique}. General agent benchmarks broaden this setting to real-world tasks involving web browsing and tool use, including information needs that are time-consuming for humans to resolve \cite{mialon2024gaia,yoran2024assistantbench}. More recent benchmarks extend the search horizon through multi-step website traversal, real-time web browsing, and information synthesis \cite{wu2025webwalker,gou2026mind2web}. BrowseComp targets persistent search and strategic query planning through hard-to-find web questions with verifiable answers \cite{wei2025browsecomp}. LiveDRBench further evaluates tasks with high search and reasoning intensity, while related benchmarks test whether agents can discover unstated search paths and synthesize verifiable findings from multiple sources \cite{java2026characterizing,song2026demystifying,paul2026deepsynth}. Controlled evaluation environments use frozen web snapshots or fixed, human-verified corpora to support reproducible comparisons between deep search agents \cite{bosse2025deepresearchbench,chen2025browsecompplus}. These benchmarks substantially broaden the evaluation of deep research, but they do not test the coordinated use of open web search and executable querying over a relational database.

\subsection{Evaluating Database Agents}

Early text-to-SQL benchmarks treated the task as mapping a natural-language question to an executable query \cite{zhong2017seq2sql,yu2018spider}. Subsequent benchmarks broadened this setting to contextual and conversational queries, larger databases, and tasks requiring database values, external knowledge, or execution efficiency \cite{yu2019sparc,yu2019cosql,li2023can}. Agent-oriented benchmarks extend the evaluation target beyond the final SQL query, with Spider 2.0 evaluating enterprise text-to-SQL involving database metadata, documentation, and project code, BIRD-Critic focusing on the diagnosis and repair of SQL issues, and BIRD-Interact evaluating dynamic interaction with users and databases \cite{lei2025spider,li2025swe,huo2026bird}. LiveSQLBench addresses another evaluation concern by continuously updating its databases and tasks to reduce benchmark contamination \cite{livesqlbench2025}. Despite supporting increasingly complex workflows, these benchmarks assume that the required evidence is contained in a predefined database and a fixed set of supporting materials, without considering settings in which additional evidence must be discovered from the open web.

\subsection{About Hybrid Benchmarks}

Real-world information-seeking tasks often require models to browse the web, use external tools, and integrate evidence from multiple sources \cite{mialon2024gaia,yoran2024assistantbench}. Evidence from heterogeneous sources can provide complementary coverage \cite{christmann2024compmix}, motivating hybrid benchmarks that evaluate source selection and cross-source reasoning rather than treating each capability in isolation. Within deep research, recent benchmarks have extended web-based information seeking across languages, specialized domains, multimodal evidence, and heterogeneous enterprise artifacts \cite{zhou2025browsecompzh,chen2025medbrowsecomp,li2025mmbrowsecomp,ning2026mcsearch,choubey2025benchmarking}. Enterprise evaluations have also begun to combine public web evidence with private knowledge bases or domain-specific internal APIs \cite{abaskohi2026drbench,ferreira2026deepresearchretail}. These settings broaden evidence access beyond the public web, but internal information is exposed through document retrieval or application-specific tools rather than executable queries over a relational database. A complementary line of hybrid question answering requires joint reasoning over structured records and textual passages \cite{chen2020hybridqa,chen2021ottqa,zhu2021tatqa}. These benchmarks demonstrate the complementary value of structured and textual evidence, but generally assume that the relevant tables and passages have been collected in advance. Related work on hybrid database querying combines database execution with language-model reasoning or knowledge, but does not require agents to discover supporting evidence through open web search \cite{li2021dual,biswal2025text2sql,zhao2025hybrid}. Recent work such as DataSTORM combines internet research with large-scale database exploration, but focuses on thesis-driven analysis and report generation rather than answer-oriented evaluation \cite{liu2026datastorm}. \OURDATA{} addresses this gap through three task types designed to capture SQL-to-Search, Search-to-SQL, and parallel coordination between search and SQL.

\end{document}